\documentclass[11pt, oneside]{article}

\usepackage[english]{babel}
\usepackage{geometry}
\usepackage{graphicx}
\usepackage{tabularx}
\usepackage{amssymb}
\usepackage{amsmath}
\usepackage{booktabs}
\usepackage{xltabular}
\usepackage[hyphens]{url}
\usepackage{hyperref}
\hypersetup{breaklinks=true,hidelinks}
\usepackage[backend=biber,style=nature]{biblatex}
\usepackage{csquotes}
\usepackage{cleveref}
\usepackage{siunitx}
\usepackage{geometry}
\usepackage{multirow}
\usepackage[dvipsnames]{xcolor}
\definecolor{tumblue}{rgb}{0,0.396078431372549,0.741176470588235}

\usepackage{xspace}
\usepackage{caption}
\usepackage{subcaption}
\usepackage{authblk}
\usepackage{mhchem}

\usepackage{marvosym}

\usepackage{sectsty}
\usepackage[acronym]{glossaries}

\sectionfont{\color{tumblue}\selectfont\sffamily}
\subsectionfont{\color{tumblue}\selectfont\sffamily}
\subsubsectionfont{\color{tumblue}\selectfont\sffamily}
\paragraphfont{\selectfont\sffamily}
\subparagraphfont{\color{gray}\selectfont\sffamily}

\definecolor{halfgray}{gray}{0.55}

\newcommand{\chembench}{ChemBench\xspace}
\newcommand{\chembenchmini}{ChemBench-Mini\xspace}

\IfFileExists{./fonts/Heliotrope-Bold.otf}{
    \usepackage{fontspec}
    \setmainfont{Heliotrope}[Extension = .otf, UprightFont = *-Regular, ItalicFont = *-Italic, BoldFont=*-Bold, Path = ./fonts/]
}{
    \usepackage{biolinum}
    
}

\usepackage{lipsum}
\usepackage[labelfont=bf]{caption}
\makenoidxglossaries 
\newacronym{llm}{LLM}{large language model}
\newacronym{ml}{ML}{machine learning}
\newacronym{agi}{AGI}{artificial general intelligence}
\newacronym{api}{API}{application programming interface}
\newacronym{mcq}{MCQ}{multiple-choice question}
\newacronym{rest}{REST}{representational state transfer}
\newacronym{orm}{ORM}{object relational mapping}
\newacronym{dai}{DAI}{daily allowed intake}
\newacronym{ghs}{GHS}{globally harmonized system of classification and labelling of chemicals}
\newacronym{who}{WHO}{World Health Organization}
\newacronym{nmr}{NMR}{nuclear magnetic resonance}
\newacronym{helm}{HELM}{Holistic Evaluation of Language Models}
\newacronym{smiles}{SMILES}{simplified molecular input line-entry system}
\newacronym{pca}{PCA}{principal component analysis}
\newacronym{iupac}{IUPAC}{International Union of Pure and Applied Chemistry}
\newacronym{json}{JSON}{JavaScript object notation}
\newacronym{rag}{RAG}{retrieval augmented generation}
\newacronym{ece}{ECE}{expected calibration error}
\newacronym{cot}{CoT}{chain of thought}
\newacronym{rlhf}{RLHF}{reinforcement learning with human feedback}

 \usepackage{microtype}

\newcommand{\oone}{o1\xspace}

\newcommand{\LlamaThreeOneSeventyBInstruct}{Llama-3.1-70B-Instruct\xspace}
\newcommand{\ClaudeThree}{Claude-3\xspace}
\newcommand{\ClaudeThreeFiveSonnet}{Claude-3.5 (Sonnet)\xspace}

\newcommand{\GPTFourO}{GPT-4o\xspace}

\newcommand{\LlamaThreeSeventyBInstruct}{Llama-3-70B-Instruct\xspace}
\newcommand{\PaperQATwo}{PaperQA2\xspace}

\newcommand{\GemmaTwoNineBIt}{Gemma-2-9B-it\xspace}

\newcommand{\LlamaThreeOneEightBInstruct}{Llama-3.1-8B-Instruct\xspace}

\newcommand{\GPTFour}{GPT-4\xspace}

\newcommand{\GPTThreeFiveTurboZeroT}{GPT-3.5 Turbo Zero-T\xspace}
\newcommand{\ClaudeTwo}{Claude-2\xspace}

\newcommand{\LlamaThreeOneFourZeroFiveBInstruct}{Llama-3.1-405B-Instruct\xspace}

\usepackage{credits}
\usepackage{orcidlink}
\usepackage{showyourwork}

\title{\textsf{Are large language models superhuman chemists?}}

\credit{Adrian~Mirza}{1,1,1,0,1,1,0,0,1,0,1,1,0,1}
\credit{Nawaf~Alampara}{1,1,1,0,1,1,0,0,1,0,1,1,0,1}
\credit{Sreekanth~Kunchapu}{1,1,1,0,1,1,0,0,1,0,1,1,0,1}
\credit{Martiño~Ríos-García}{1,1,1,0,1,1,0,0,1,0,1,1,0,1}

\credit{Benedict~Emoekabu}{0,1,0,0,1,0,0,0,0,0,0,0,0,1}
\credit{Aswanth~Krishnan}{0,0,0,0,0,0,0,0,1,0,0,0,0,1}
\credit{Tanya~Gupta}{0,1,0,0,1,0,0,0,0,0,0,0,0,1}
\credit{Mara~Schilling-Wilhelmi}{0,1,0,0,1,0,0,0,0,0,0,0,0,1}
\credit{Macjonathan~Okereke}{0,1,0,0,1,0,0,0,0,0,0,0,0,1}

\credit{Anagha Aneesh}{0,1,0,0,0,0,0,0,0,0,0,0,0,1}
\credit{Mehrdad Asgari}{0,1,0,0,0,0,0,0,0,0,0,0,0,1}
\credit{Juliane Eberhardt}{0,1,0,0,0,0,0,0,0,0,0,0,0,1}
\credit{Amir~Mohammad~Elahi}{0,1,0,0,0,0,0,0,0,0,0,0,0,1}
\credit{Hani~M.~Elbeheiry~}{0,1,0,0,0,0,0,0,0,0,0,0,0,1}
\credit{María~Victoria~Gil}{0,1,0,0,1,0,0,0,0,0,0,0,0,1}
\credit{Christina Glaubitz}{0,1,0,0,0,0,0,0,0,0,0,0,0,1}
\credit{Maximilian Greiner}{0,1,0,0,0,0,0,0,0,0,0,0,0,1}
\credit{Caroline T. Holick}{0,1,0,0,0,0,0,0,0,0,0,0,0,1}
\credit{Tim Hoffmann}{0,1,0,0,0,0,0,0,0,0,0,0,0,1}
\credit{Abdelrahman~Ibrahim}{0,1,0,0,0,0,0,0,0,0,0,0,0,1}
\credit{Lea~C.~Klepsch}{0,1,0,0,0,0,0,0,0,0,0,0,0,1}
\credit{Yannik Köster}{0,1,0,0,0,0,0,0,0,0,0,0,0,1}
\credit{Fabian Alexander Kreth}{0,1,0,0,0,0,0,0,0,0,0,0,0,1}
\credit{Jakob~Meyer}{0,1,0,0,0,0,0,0,0,0,0,0,0,1}
\credit{Santiago~Miret}{0,1,0,0,0,0,0,0,0,0,0,0,0,1}
\credit{Jan Matthias Peschel}{0,1,0,0,0,0,0,0,0,0,0,0,0,1}
\credit{Michael Ringleb}{0,1,0,0,0,0,0,0,0,0,0,0,0,1}
\credit{Nicole~C.~Roesner}{0,1,0,0,0,0,0,0,0,0,0,0,0,1}
\credit{Johanna Schreiber}{0,1,0,0,0,0,0,0,0,0,0,0,0,1}
\credit{Ulrich~S.~Schubert}{0,0,0,1,0,0,0,0,0,0,0,0,0,1}
\credit{Leanne M. Stafast}{0,1,0,0,0,0,0,0,0,0,0,0,0,1}
\credit{Dinga Wonanke}{0,1,0,0,0,0,0,0,0,0,0,0,0,1}

\credit{Michael~Pieler}{1,0,0,1,0,1,0,1,0,0,0,0,0,1}
\credit{Philippe~Schwaller}{1,1,0,1,1,1,0,0,0,1,0,0,0,1}
\credit{Kevin Maik Jablonka}{1,1,1,1,1,1,1,1,1,1,1,1,1,1}

\author[1,2,$\star$]{Adrian~Mirza~\orcidlink{0000-0003-4033-4235}}
\author[1,$\star$]{Nawaf~Alampara~\orcidlink{0009-0001-7697-7315}}
\author[1,$\star$]{Sreekanth~Kunchapu~\orcidlink{0009-0003-5752-0154}}
\author[1,3 $\star$]{Martiño~Ríos-García~\orcidlink{0000-0003-1507-4048}}
\author[ ]{Benedict~Emoekabu}
\author[4]{Aswanth~Krishnan~\orcidlink{0009-0008-2703-5613}}
\author[5,6]{Tanya~Gupta~\orcidlink{0009-0001-9523-3290}}
\author[1]{Mara~Schilling-Wilhelmi~\orcidlink{0009-0007-4392-5918}}
\author[1]{Macjonathan~Okereke~\orcidlink{0009-0007-1013-0502}}

\author[1]{Anagha~Aneesh~\orcidlink{0009-0001-0275-2586}}
\author[7]{Mehrdad~Asgari~\orcidlink{0000-0002-5427-1610}}
\author[8]{Juliane~Eberhardt~\orcidlink{0009-0000-3991-0704}}
\author[9]{Amir~Mohammad~Elahi~\orcidlink{0009-0001-5907-101X}}
\author[10]{Hani~M.~Elbeheiry~\orcidlink{0000-0002-5205-2852}}
\author[3]{María~Victoria~Gil~\orcidlink{0000-0002-2258-3011}}
\author[ ]{Christina~Glaubitz~\orcidlink{0000-0003-3412-0548}}
\author[1]{Maximilian~Greiner}
\author[1,14]{Caroline~T.~Holick~\orcidlink{0009-0000-1724-2725}}
\author[1, 14]{Tim~Hoffmann~\orcidlink{0009-0004-0230-6115}}
\author[1]{Abdelrahman~Ibrahim~\orcidlink{0009-0003-1460-4710}}
\author[1, 14]{Lea~C.~Klepsch~\orcidlink{0009-0009-3849-1670}}
\author[1]{Yannik~Köster~\orcidlink{0000-0002-9125-3067}}
\author[11, 12]{Fabian~Alexander~Kreth~\orcidlink{0000-0002-5968-8706}}
\author[1]{Jakob~Meyer}
\author[13]{Santiago~Miret~\orcidlink{0000-0002-4787-2757}}
\author[1]{Jan~Matthias~Peschel~\orcidlink{0009-0002-4787-2757}}
\author[1, 14]{Michael~Ringleb~\orcidlink{0000-0002-7320-8529}}
\author[1, 14]{Nicole~Roesner~\orcidlink{0000-0002-5133-775X}}

\author[1, 14]{Johanna~Schreiber~\orcidlink{0009-0000-0991-8967}}
\author[1,2, 10, 14]{Ulrich~S.~Schubert~\orcidlink{0000-0003-4978-4670}}
\author[1, 14]{Leanne~M.~Stafast~\orcidlink{0009-0008-5604-261X}}
\author[15]{Dinga~Wonanke~\orcidlink{0000-0002-9066-2715}}

\author[16,17]{Michael~Pieler~\orcidlink{0000-0001-9186-7045}}
\author[5, 6]{Philippe~Schwaller~\orcidlink{0000-0003-3046-6576}}
\author[1,2, 11, 14, \Letter]{Kevin~Maik~Jablonka~\orcidlink{0000-0003-4894-4660}}

\affil[1]{Laboratory of Organic and Macromolecular Chemistry (IOMC), Friedrich Schiller University Jena, Humboldtstrasse 10, 07743 Jena, Germany}
\affil[2]{Helmholtz Institute for Polymers in Energy Applications Jena (HIPOLE Jena), Lessingstrasse 12-14, 07743 Jena, Germany}

\affil[3]{Institute of Carbon Science and Technology (INCAR), CSIC, Francisco Pintado Fe 26, 33011 Oviedo, Spain}
\affil[4]{QpiVolta Technologies Pvt Ltd}
\affil[5]{Laboratory of Artificial Chemical Intelligence (LIAC), Institut des Sciences et Ing\'{e}nierie Chimiques, Ecole Polytechnique F\'{e}d\'{e}rale de Lausanne (EPFL), Lausanne, Switzerland}
\affil[6]{National Centre of Competence in Research (NCCR) Catalysis, Ecole Polytechnique F\'{e}d\'{e}rale de Lausanne (EPFL), Lausanne, Switzerland}
\affil[7]{Department of Chemical Engineering \& Biotechnology, University of Cambridge, Philippa Fawcett Drive, Cambridge CB3 0AS, United Kingdom}
\affil[8]{Macromolecular Chemistry, University of Bayreuth, 95447 Bayreuth, Germany}
\affil[9]{Laboratory of Molecular Simulation (LSMO), Institut des Sciences et Ing\'{e}nierie Chimiques, Ecole Polytechnique F\'{e}d\'{e}rale de Lausanne (EPFL), Sion, Switzerland}
\affil[10]{Institute for Inorganic and Analytical Chemistry (IAAC), Friedrich Schiller University Jena, Humboldtstrasse 8, 07743 Jena, Germany}

\affil[11]{Center for Energy and Environmental Chemistry Jena (CEEC Jena), Friedrich Schiller University Jena, Philosophenweg 7a, 07743 Jena, Germany}
\affil[12]{Institute for Technical Chemistry and Environmental Chemistry (ITUC), Friedrich Schiller University Jena, Philosophenweg 7a, 07743 Jena, Germany}
\affil[13]{Intel Labs}
\affil[14]{Jena Center for Soft Matter (JCSM), Friedrich Schiller University Jena, Philosophenweg 7, 07743 Jena, Germany}

\affil[15]{Theoretical Chemistry, Technische Universität Dresden, Dresden 01062, Germany}
\affil[16]{OpenBioML.org}
\affil[17]{Stability.AI}

\affil[\Letter]{\texttt{mail@kjablonka.com}}
\affil[$\star$]{These authors contributed equally.}

\begin{document}
\maketitle

\clearpage
\begin{abstract}
Large language models (LLMs) have gained widespread interest due to their ability to process human language and perform tasks on which they have not been explicitly trained.

However, we possess only a limited systematic understanding of the chemical capabilities of LLMs, which would be required to improve models and mitigate potential harm.
Here, we introduce \enquote{\chembench,} an automated framework for evaluating the chemical knowledge and reasoning abilities of state-of-the-art LLMs against the expertise of chemists.

We curated more than 2,700 question-answer pairs, evaluated leading open- and closed-source LLMs, and found that the best models outperformed the best human chemists in our study on average.
However, the models struggle with some basic tasks and provide overconfident predictions.

These findings reveal LLMs' impressive chemical capabilities while emphasizing the need for further research to improve their safety and usefulness. They also suggest adapting chemistry education and show the value of benchmarking frameworks for evaluating LLMs in specific domains.
\end{abstract}

\clearpage
\begin{refsection}
\section{Introduction}
\Glspl{llm} are \gls{ml} models trained on massive amounts of text to complete sentences.
Aggressive scaling of these models has led to a rapid increase in their capabilities,\autocite{brown2020language,zhong2024benchmarking} with the leading models now being able to pass the United States Medical Licensing Examination\autocite{kung2023performance} or other professional licensing exams.
They also have been shown to design and autonomously perform chemical reactions when augmented with external tools such as web search and synthesis planners.\autocite{openai2024gpt4, Boiko_2023, bran2023chemcrow, darvish2024organa}
While some see \enquote{sparks of \gls{agi}} in them,\autocite{bubeck2023sparks} others consider them as \enquote{stochastic parrots}---i.e., systems that only regurgitate what they have been trained on\autocite{bender2021dangers} and that show inherent limitations due to the way they are trained.\autocite{mccoy2023embersautoregressionunderstandinglarge}
Nevertheless, the promise of these models is that they have shown the ability to solve a wide variety of tasks they have not been explicitly trained on.\autocite{bommasani2021opportunities, anderljung2023frontier, ai4science2023impact}

Chemists and materials scientists have quickly caught on to the mounting attention given to \glspl{llm}, with some voices even suggesting that \enquote{the future of chemistry is language.}\autocite{White_2023}
This statement is motivated by a growing number of reports that use \glspl{llm} to predict properties of molecules or materials,\autocite{jablonka202314, jablonka2024leveraging, xie2024fine, liao2024words, zhang2024chemllm, zhong2024benchmarking}  optimize reactions,\autocite{ramos2023bayesian, kristiadi2024sober}  generate materials,\autocite{rubungo2023llm, flam2023language, gruver2024fine, alampara2024mattextlanguagemodelsneed} extract information,\autocite{Patiny_2023, Dagdelen_2024, Zheng_2024, lala2023paperqa, caufield2023structured, gupta2022discomat, schillingwilhelmi2024textinsightlargelanguage, skarlinski2024language} or to even prototype systems that can autonomously perform experiments in the physical world based on commands provided in natural language.\autocite{bran2023chemcrow, Boiko_2023, darvish2024organa}

In addition, since a lot---if not most---of the information about chemistry is currently stored and communicated in text, there is a strong reason to believe that there is still a lot of untapped potential in \glspl{llm} for chemistry and materials science.\autocite{miret2024llms}
For instance, most insights in chemical research do not directly originate from data stored in databases but rather from the scientists interpreting the data.
Many of these insights are in the form of text in scientific publications.
Thus, operating on such texts might be our best way of unlocking these insights and learning from them.
This might ultimately lead to general copilot systems for chemists that can provide answers to questions or even suggest new experiments based on vastly more information than a human could ever read.

However, the rapid increase in capabilities of chemical \gls{ml} models led (even before the recent interest in \glspl{llm}) to concerns about the potential for dual use of these technologies, e.g., for the design of chemical weapons.\autocite{gopal2023releasing, ganguli2022red, Urbina_2022, campbell2023censoring, moulange2023towards, urbina2022teachable}
To some extent, this is not surprising as any technology that, for instance, is used to design non-toxic molecules can also be used inversely to predict toxic ones (even though the synthesis would still require access to controlled physical resources and facilities).
Still, it is essential to realize that the user base of \glspl{llm} is broader than that of chemistry and materials science experts who can critically reflect on every output these models produce.
For example, many students frequently consult these tools---perhaps even to prepare chemical experiments.\autocite{Intelligent.com_2023}
This also applies to users from the general public, who might consider using \glspl{llm} to answer questions about the safety of chemicals.
Thus, for some users, misleading information---especially about safety-related aspects---might lead to harmful outcomes.
However, even for experts, chemical knowledge and reasoning capabilities are essential as they will determine the capabilities and limitations of their models in their work, e.g., in copilot systems for chemists.
Unfortunately, apart from exploratory reports such as by prompting leading models with various scientific questions,\autocite{ai4science2023impact} there is little systematic evidence on how \glspl{llm} perform compared to expert (human) chemists.

Thus, to better understand what \glspl{llm} can do for the chemical sciences and where they might be improved with further developments, evaluation frameworks are needed to allow us to measure progress and mitigate potential harms systematically.
For the development of \glspl{llm}, evaluation is currently primarily performed via standardized benchmark suites such as BigBench\autocite{srivastava2022beyond} or the LM Eval Harness.\autocite{eval-harness}
Among 204 tasks (such as linguistic puzzles), the former contains only two tasks classified as \enquote{chemistry related}, whereas the latter contains no specific chemistry tasks.
Due to the lack of widely accepted standard benchmarks, the developers of chemical language models\autocite{jablonka2024leveraging, guo2023large, ahmad2022chemberta2, Cai_2024, frey2023neural} frequently utilize language-interfaced\autocite{dinh2022lift} tabular datasets such as the ones reported in MoleculeNet,\autocite{wu2018moleculenet} Therapeutic Data Commons\autocite{huang2021therapeutics} or MatBench.\autocite{Dunn_2020}
In these cases, the models are evaluated on predicting very specific properties of molecules (e.g., solubility, toxicity, melting temperature or reactivity) or on predicting the outcome of specific chemical reactions.
This, however, only gives a very limited view of the general chemical capabilities of the models.

While some benchmarks based on university entrance exams\autocite{Zaki_2024, arora2023llms} or automatic text mining\autocite{song2023honeybee, wei2021chemistryqa, song-etal-2023-matsci} have been proposed, none of them have been widely accepted.
This is likely because they cannot automatically be used with black box (or tool-augmented) systems, do not cover a wide range of topics and skills, or are not carefully validated by experts.
On top of that, the existing benchmarks are not designed to be used with models that support special treatment of molecules or equations and do not provide insights on how the models compare relative to experts \autocite{wu2018moleculenet}.

\begin{figure}
    \includegraphics[width=\textwidth]{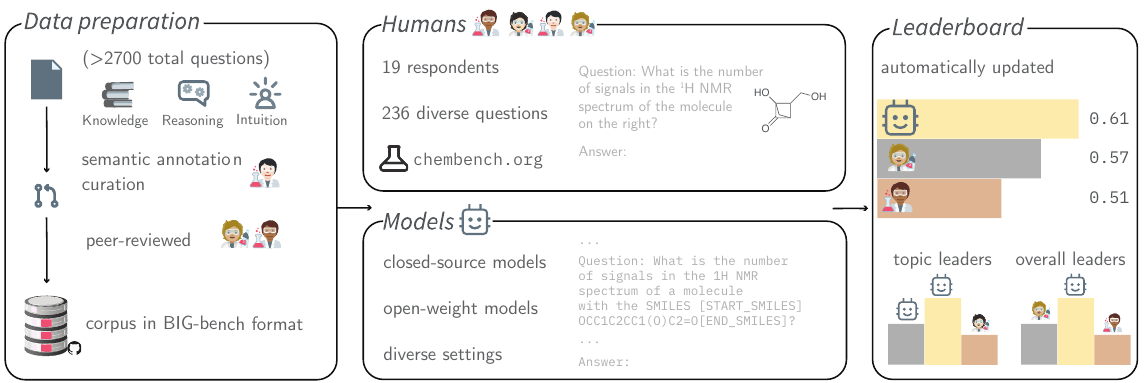}
    \caption{\textbf{Overview of the \chembench framework.} The figure shows the different components of the \chembench framework.
    The framework's foundation is the benchmark corpus comprising thousands of questions and answers that we manually or semi-automatically compiled from various sources (see \Cref{sec:curation}).
    Questions are classified based on topics, required skills (reasoning, calculation, knowledge, intuition), and difficulty levels.
    We then used this corpus to evaluate the performance of various models and tool-augmented systems using a custom framework. To provide a baseline, we built a web application that we used to survey experts in chemistry.
    The results of the evaluations are then compiled in publicly accessible leaderboards (\Cref{sec:leaderboard}), which we propose as a foundation for evaluating future models.
    }
    \label{fig:overview_figure}
\end{figure}

In this work, we report a novel benchmarking framework  (\Cref{fig:overview_figure}), which we call \chembench, and use it to reveal limitations of current frontier models for use in the chemical sciences.
Our benchmark consists of %
  2788 \unskip\label{output/total_number_of_questions.txt}\unskip%
\xspace question-answer pairs compiled from diverse sources (%
  1039 \unskip\label{output/manually_generated.txt}\unskip%
\xspace manually generated, and %
  1749 \unskip\label{output/automatically_generated.txt}\unskip%
\xspace semi-automatically generated).
Our corpus measures reasoning, knowledge and intuition across a large fraction of the topics taught in undergraduate and graduate chemistry curricula. It can be used to evaluate any system that can return text (i.e., including tool-augmented systems).

To contextualize the scores, we also surveyed %
  19 \unskip\label{output/number_experts.txt}\unskip%
 experts in chemistry on a subset of the benchmark corpus to be able to compare the performance of current frontier models with (human) chemists of different specializations. In parts of the survey, the volunteers were also allowed to use tools such as web search to create a realistic setting.

\section{Results and Discussion}

\subsection{Benchmark corpus}

\begin{figure}[!htb]
    \centering
    \includegraphics[width=\textwidth]{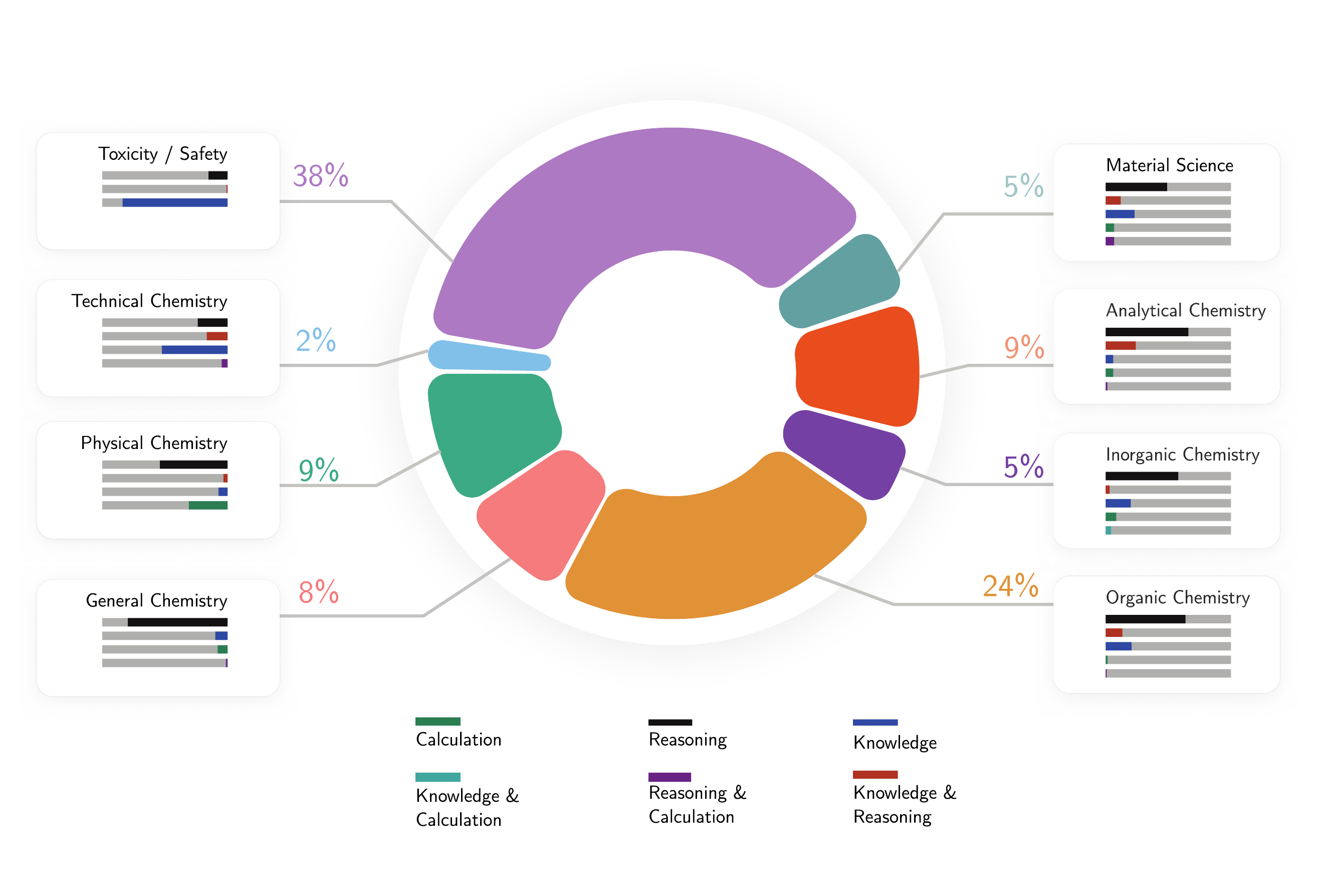}
    \caption{\textbf{Distribution of topics and required skills.} This circular plot illustrates the distribution of questions across various chemistry topics, along with the primary skills required to address them. The topics were manually classified, showing a varied representation across different aspects of chemistry. Each topic is associated with a combination of three key skills: Calculation, Reasoning, and Knowledge, as indicated by the colored bars. \chembench samples diverse topics and diverse skills, setting a high bar for \glspl{llm} to demonstrate human-competitive performance across a wide range of chemistry tasks.}
    \label{fig:corpus}
\end{figure}

To compile our benchmark corpus, we utilized a broad list of sources (see \Cref{sec:curation}), ranging from completely novel, manually crafted questions over university exams to semi-automatically generated questions based on curated subsets of data in chemical databases.
For quality assurance, all questions have been reviewed by at least two scientists in addition to the original curator and automated checks.
Importantly, our large pool of questions encompasses a wide range of topics and question types (\Cref{fig:corpus}). The topics range from general chemistry to more specialized fields such as inorganic, analytical or technical chemistry.
We also classify the questions based on what skills are required to answer them. Here, we distinguish between questions that require knowledge, reasoning, calculation, intuition, or a combination of these.
Moreover, the annotator also classifies the questions by difficulty to allow for a more nuanced evaluation of the models' capabilities.

While many existing benchmarks are designed around \gls{mcq}, this does not reflect the reality of chemistry education and research.
For this reason, \chembench samples both \gls{mcq} and open-ended questions (%
  2544 \unskip\label{output/mcq_questions.txt}\unskip%
 \gls{mcq} questions and %
  244 \unskip\label{output/non_mcq_questions.txt}\unskip%
 open-ended questions). In addition, \chembench samples different skills on various difficulty levels: From basic knowledge questions (as knowledge underpins reasoning processes\autocite{hu2024towards,bloom1956taxonomy}) to complex reasoning tasks (such as finding out which ions are in a sample given a description of observations). We also include questions about chemical intuition, as showing human-aligned preferences is relevant for applications such as hypothesis generation or optimization tasks.\autocite{zhang2024omniopenendednessmodelshuman}

\paragraph{\chembenchmini}
It is important to note that a smaller subset of the corpus might be more practical for routine evaluations.\autocite{polo2024tinybenchmarks}
For instance,~\textcite{liang2023holistic} report costs of more than \$10,000 for \gls{api} calls for a single evaluation on the widely used \gls{helm} benchmark.
To address this, we also provide a subset (\chembenchmini, %
  236 \unskip\label{output/num_human_answered_questions.txt}\unskip%
 questions) of the corpus that was curated to be a diverse and representative subset of the full corpus. While it is impossible to comprehensively represent the full corpus in a subset, we aimed to include a maximally diverse set of questions and a more balanced distribution of topics and skills (see \Cref{sec:subset-selection} for details on the curation process).
Our human volunteers answered all the questions in this subset.

\subsection{Model evaluation}

\paragraph{Benchmark suite design} Because the text used in scientific settings differs from typical natural language, many models have been developed that deal with such text in a particular way.
For instance, the Galactica model\autocite{taylor2022galactica} uses special encoding procedures for molecules and equations.
Current benchmarking suites, however, do not account for such special treatment of scientific information.
To address this, \chembench encodes the semantic meaning of various parts (e.g., chemicals, units, equations) of the question or answer.
For instance, molecules represented in \gls{smiles} are enclosed in \texttt{[START\_SMILES][\textbackslash END\_SMILES]} tags.
This allows the model to treat the \gls{smiles} string differently from other text.
\chembench can seamlessly handle such special treatment in an easily extensible way because the questions are stored in an annotated format.

Since many widely utilized \gls{llm} systems only provide access to text completions (and not the raw model outputs), \chembench is designed to operate on text completions.
This is also important given the growing number of tool-augmented systems that are deemed essential for building chemical copilot systems.
Such systems can augment the capabilities of \glspl{llm} through the use of external tools such as search \glspl{api} or code executors.\autocite{schick2024toolformer, karpas2022mrkl, yao2022react}
In those cases, the \gls{llm} that returns the probabilities for various tokens, i.e., text fragments, is only a part of the whole system, and it is not clear how to interpret the probabilities in the context of the whole system.
The text completions, however, are the system's final outputs, which would also be used in a real-world application.
Hence, we use them for our evaluations.\autocite{xiong2023llms}

\begin{figure}[!h]
    \centering
    \includegraphics{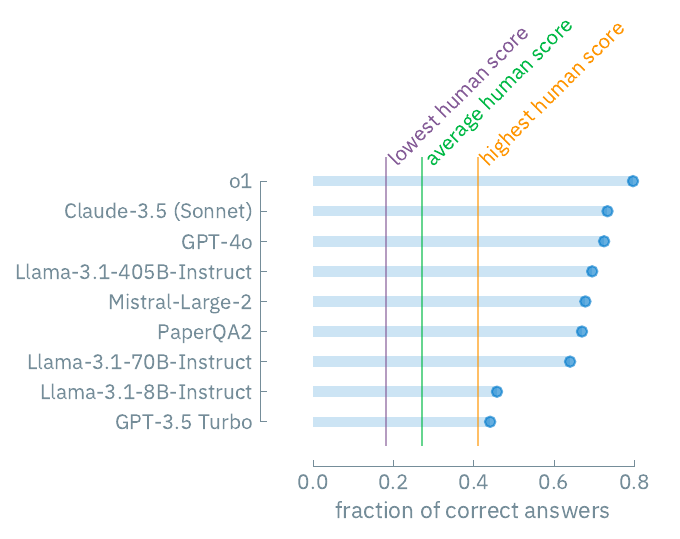}
    \caption{\textbf{Performance of models and humans on \chembenchmini.} The figure shows the percentage of questions that the models answered correctly. We use horizontal bars to indicate the performance of various models and highlight statistics of the human performance.
    The evaluation we use here is very strict as it only considers a question answered correctly or incorrectly, partially correct answers are also considered incorrect.
    \Cref{fig:barplot_all_correct_all_questions} provides an overview of the performance of various models on the entire corpus.
    PaperQA2\autocite{skarlinski2024language} is an agentic system that can also search the literature to obtain an answer. We find that the best models outperform all humans in our study when averaged over all questions (even though humans had access to tools such as web search and ChemDraw for a subset of the questions).
    }
    \label{fig:human_vs_models_bar}
    \script{plot_overview_performance_plot.py}
\end{figure}

\paragraph{Overall system performance}
To understand the current capabilities of \glspl{llm} in the chemical sciences, we evaluated a wide range of leading models\autocite{Huggingface} on the \chembench corpus, including systems augmented with external tools.
An overview of the results of this evaluation is shown in \Cref{fig:human_vs_models_bar} (all results can be found in \Cref{tab:performance_table} and \Cref{tab:performance_table_human_subset}).
In this figure, we show the percentage of questions that the models answered correctly.
Moreover, we show the worst, best, and average performance of the experts in our study, which we obtained via a custom web application (\url{chembench.org}) that we used to survey the experts.
Remarkably, the figure shows that the leading \gls{llm}, \oone, outperforms the best human in our study in this overall metric by almost a factor of two.
Many other models also outperform the average human performance.
Interestingly, \LlamaThreeOneFourZeroFiveBInstruct shows performance that is close to the leading proprietary models, indicating that new open-source models can be competitive with the best proprietary models also in chemical settings.

Notably, we find that models are still limited in their ability to answer knowledge-intensive questions (\Cref{tab:performance_table_human_subset}); that is, they did not memorize the relevant facts. Our results indicate that this is not a limitation that could be overcome by simple application of \gls{rag} systems such as \PaperQATwo. This is likely because the required knowledge cannot easily be accessed via papers (which is the only external knowledge \PaperQATwo has access to) but rather by lookup in specialized databases (e.g., PubChem, Gestis), which also the humans in our study used to answer such questions (\Cref{fig:tool_use}).
This indicates that there is still room for improving chemical \glspl{llm} by training them on more specialized data sources or integrating them with specialized databases.

In addition, our analysis shows that the performance of models is correlated with their size (see \Cref{fig:model_size_plot}). This is in line with observations in other domains but also indicates that chemical \glspl{llm} could, to some extent, be further improved by scaling them up.

\begin{figure}[!h]
    \centering
    \includegraphics[width=\textwidth]{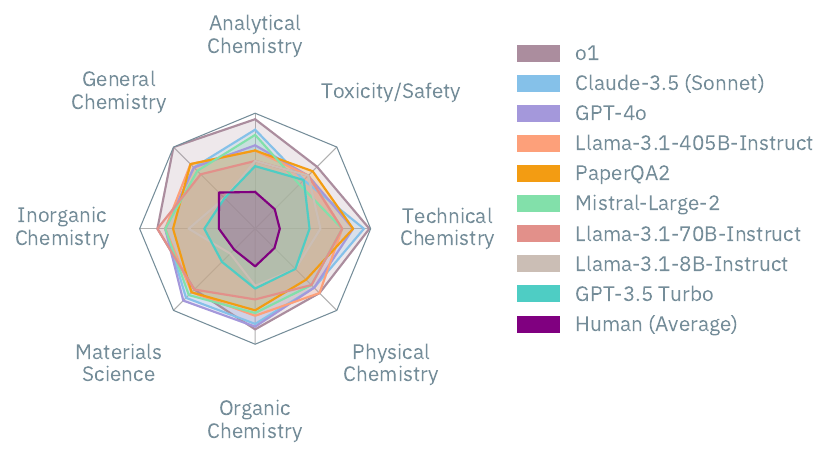}
    \caption{\textbf{Performance of the models and humans on the different topics on \chembenchmini.} The radar plot shows the performance of the models and humans on the different topics of \chembenchmini. The performance is measured as the fraction of questions that were answered correctly by the models.
    The best score for every dimension is one (all questions answered correctly), and the worst is zero (no question answered correctly). A larger colored area indicates a better performance.
    This figure shows the performance on \chembenchmini. The performance of models on the entire corpus is shown in \Cref{fig:barplot_all_correct_all_questions}.
    }
    \label{fig:all_questions_models_completely_correct_radar_human}
    \script{analyze_model_reports.py}
\end{figure}

\paragraph{Performance per topic} To obtain a more detailed understanding of the performance of the models, we also analyzed the performance of the models in different subfields of the chemical sciences.
For this analysis, we defined a set of topics (see \Cref{sec:meth-topic}) and classified all questions in the \chembench corpus into these topics.
We then computed the percentage of questions the models or experts answered correctly for each topic and show them in \Cref{fig:all_questions_models_completely_correct_radar_human}.
In this spider chart, the worst score for every dimension is zero (no question answered correctly), and the best score is one (all questions answered correctly). Thus, a larger colored area indicates a better performance.

One can observe that this performance varies across models and topics.
While general and technical receive relatively high scores for many models, this is not the case for topics such as toxicity and safety or analytical chemistry.

In the subfield of analytical chemistry, the prediction of the number of signals observable in a \gls{nmr} spectrum proved difficult even for the best models (e.g., %
  22 \unskip\label{output/subset_scores/is_number_nmr_peaks_o1.txt}\unskip%
 percent correct answers for \oone).
Importantly, while the human experts are given a drawing of the compounds, the models are only shown the \gls{smiles} string of a compound and have to use this to reason about the symmetry of the compound (i.e., to identify the number of diasterotopically distinct protons, which requires \emph{reasoning} about the topology and structure of a molecule).

 These findings also shine an interesting light on the value of textbook-inspired questions.
 A subset of the questions in \chembench are based on textbooks targeted at undergraduate students.
 On those questions, the models tend to perform better than on some of our semi-automatically constructed tasks (see \Cref{fig:performance_per_topic}).
 For instance, while the overall performance in the chemical safety topic is low, the models would pass the certification exam according to the German Chemical Prohibition Ordinance based on a subset of questions we sampled from the corresponding question bank (e.g., %
  71 \unskip\label{output/subset_scores/is_gfk_gpt-4.txt}\unskip%
\% correct answers for \GPTFour, %
  61 \unskip\label{output/subset_scores/is_gfk_Claude-3.5__Sonnet_.txt}\unskip%
\% for \ClaudeThreeFiveSonnet, and %
  3 \unskip\label{output/human_subset_scores/is_gfk.txt}\unskip%
\% for the human experts).
 While those findings are impacted by the subset of questions we sampled, the results still highlight that good performance on such question bank or textbook questions does not necessarily translate to good performance on other questions that require more reasoning or are further away from the training corpus.\autocite{mccoy2023embersautoregressionunderstandinglarge} The findings also underline that such exams might have been a good surrogate for the general performance of skills for humans, but their applicability in the face of systems that can consume vast amounts of data is up for debate.

 We also gain insight into the models' struggles with chemical reasoning tasks by examining their performance as a function of molecular descriptors.
 If the model would answer questions after reasoning about the structures, one would expect the performance to depend on the complexity of the molecules.
 However, we find that the models' performance does not correlate with complexity indicators (see \Cref{sec:molecular_features}).
 This indicates that the models may not be able to reason about the structures of the molecules (in the way one might expect) but instead rely on the proximity of the molecules to the training data.\autocite{mccoy2023embersautoregressionunderstandinglarge}

It is important to note that the model performance for some topics, however, is slightly underestimated in the current evaluation.
 This is because models provided via \glspl{api} typically have safety mechanisms that prevent them from providing answers that the provider deems unsafe.
 For instance, models might refuse to provide answers about cyanides. Statistics of the frequency of such refusals are shown in \Cref{tab:refusal_counts_and_parsing}.
 To overcome this, direct access to the model weights would be required, and we strive to collaborate with the developers of frontier models to overcome this limitation in the future.
 This is facilitated by the tooling \chembench provides, thanks to which contributors can automatically add new models in an open science fashion.

\paragraph{Judging chemical preference}

One interesting finding of recent research is that foundation models can judge interestingness or human preferences in some domains.\autocite{zhang2024omniopenendednessmodelshuman, Argyle_2023}
If models could do so for chemical compounds, this would open opportunities for novel optimization approaches.
Such open-ended tasks, however, depend on an external observer defining what interestingness is.\autocite{hughes2024openendednessessentialartificialsuperhuman}
Here, we posed models the same question~\textcite{Choung_2023} asked chemists at a drug company: \enquote{Which of the two compounds do you prefer?} (in the context of an early virtual screening campaign setting, see \Cref{tab:chembench_corpus_cognitive} for an example).
Despite chemists demonstrating a reasonable level of interrater agreement, our models largely fail to align with expert chemists' preferences. Their performance is often indistinguishable from random guessing, even though these same models excel in other tasks in \chembench (\Cref{tab:performance_table_human_subset}).
This indicates that using preference tuning for chemical settings is a promising approach to explore in future research.

\paragraph{Confidence estimates} One might wonder whether the models can estimate if they can answer a question correctly.
If they could do so, incorrect answers would be less problematic.

To investigate this, we prompted\autocite{xiong2023llms} some of the top-performing models to estimate, on an ordinal scale, their confidence in their ability to answer the question correctly (see \Cref{sec:confidence_estimates} for details on the methodology and comparison to logit-based approaches).

 \begin{figure}[!h]
     \centering
     \includegraphics[width=\textwidth]{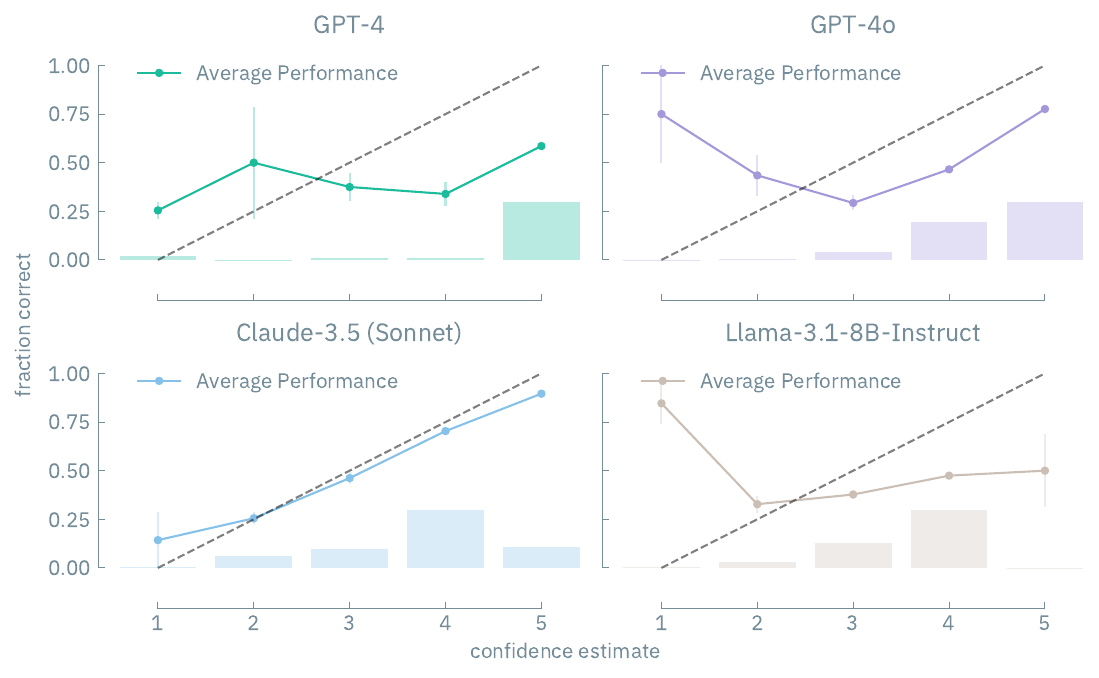}
     \caption{\textbf{Reliability and distribution of confidence estimates.} For this analysis, we used verbalized confidence estimates from the model. We prompted the models to return a confidence score on an ordinal scale to obtain those estimates.
     The line plot shows the average fraction of correctly answered questions for each confidence level. The bar plot shows the distribution of confidence estimates.
     A confidence estimate would be well-calibrated if the average fraction of correctly answered questions increases with the confidence level.      The dashed black line indicates this ideal behavior, which would be monotonically increasing correctness with higher
     levels of confidence.
     We find that most models are not well-calibrated and provide misleading confidence estimates.
     }
     \label{fig:confidence_vs_performance}
     \script{confidence_estimate.py}
 \end{figure}

 In \Cref{fig:confidence_vs_performance}, we show that for some models, there is no significant correlation between the estimated difficulty and whether the models answered the question correctly or not.
 For applications in which humans might rely on the models to provide answers with trustworthy uncertainty estimates, this is a concerning observation highlighting the need for critical reasoning in the interpretation of the model's outputs.\autocite{Li_2023, miret2024llms}
 For example, for the questions about the safety profile of compounds, \GPTFour reported a confidence of %
  1.0 \unskip\label{output/model_confidence_performance/gpt-4_is_pictograms_average_confidence_correct_overall.txt}\unskip%
 (on a scale of 1--5) for the %
  1 \unskip\label{output/model_confidence_performance/gpt-4_is_pictograms_num_correct_overall.txt}\unskip%
 questions it answered correctly and %
  4.0 \unskip\label{output/model_confidence_performance/gpt-4_is_pictograms_average_confidence_incorrect_overall.txt}\unskip%
 for the %
  6 \unskip\label{output/model_confidence_performance/gpt-4_is_pictograms_num_incorrect_overall.txt}\unskip%
 questions it answered incorrectly.
 While, on average, the verbalized confidence estimates from Claude 3.5 seem better calibrated (\Cref{fig:confidence_vs_performance}), they are still misleading in some cases.
 For example, for the questions about the \gls{ghs} pictograms Claude 3.5 returns an average score of %
  2.0 \unskip\label{output/model_confidence_performance/claude3_is_pictograms_average_confidence_correct_overall.txt}\unskip%
 for correct answers and %
  1.83 \unskip\label{output/model_confidence_performance/claude3_is_pictograms_average_confidence_incorrect_overall.txt}\unskip%
 for incorrect answers.

\section{Conclusions}
On the one hand, our findings underline the impressive capabilities of \glspl{llm} in the chemical sciences: Leading models outperform domain experts in specific chemistry questions on many topics.
On the other hand, there are still striking limitations.
For very relevant topics, the answers that models provide are wrong.
On top of that, many models are not able to reliably estimate their own limitations.
Yet, the success of the models in our evaluations perhaps also reveals more about the limitations of the questions we use to evaluate models---and chemists---than about the models themselves.
For instance, while models perform well on many textbook questions, they struggle with questions requiring more reasoning about chemical structures (e.g., number of isomers or \gls{nmr} peaks).
Given that the models outperformed the average human in our study, we need to rethink how we teach and examine chemistry.
Critical reasoning is increasingly essential, and rote solving of problems or memorization of facts is a domain in which \glspl{llm} will continue to outperform humans (when trained on the right training corpus).

Our findings also highlight the nuanced trade-off between breadth and depth of evaluation frameworks.
The analysis of model performance on different topics shows that models' performance varies widely across the subfields they are tested on.
However, even within a topic, the performance of models can vary widely depending on the type of question and the reasoning required to answer it.

The current evaluation frameworks for chemical \glspl{llm} are primarily designed to measure the performance of the models on specific property prediction tasks.
They cannot be used to evaluate reasoning or systems built for scientific applications.
Thus, we had little understanding of the capabilities of \glspl{llm} in the chemical sciences.
Our work shows that carefully curated benchmarks can provide a more nuanced understanding of the capabilities of \glspl{llm} in the chemical sciences.
Importantly, our findings also illustrate that more focus is required in developing better human-model interaction frameworks, given that models cannot estimate their limitations.

While our findings indicate many areas for further improvement of \gls{llm}-based systems, such as for agents (more discussion in \Cref{sec:react-environment}), it is also important to realize that clearly defined metrics have been the key to the progress of many fields of \gls{ml}, such as computer vision.
Although current systems might be far from reasoning like a chemist, our \chembench framework will be a stepping stone for developing systems that might come closer to this goal.

\clearpage

\section{Methods}

\subsection{Curation workflow}\label{sec:curation}
\begin{figure}[!h]
    \includegraphics[width = \textwidth]{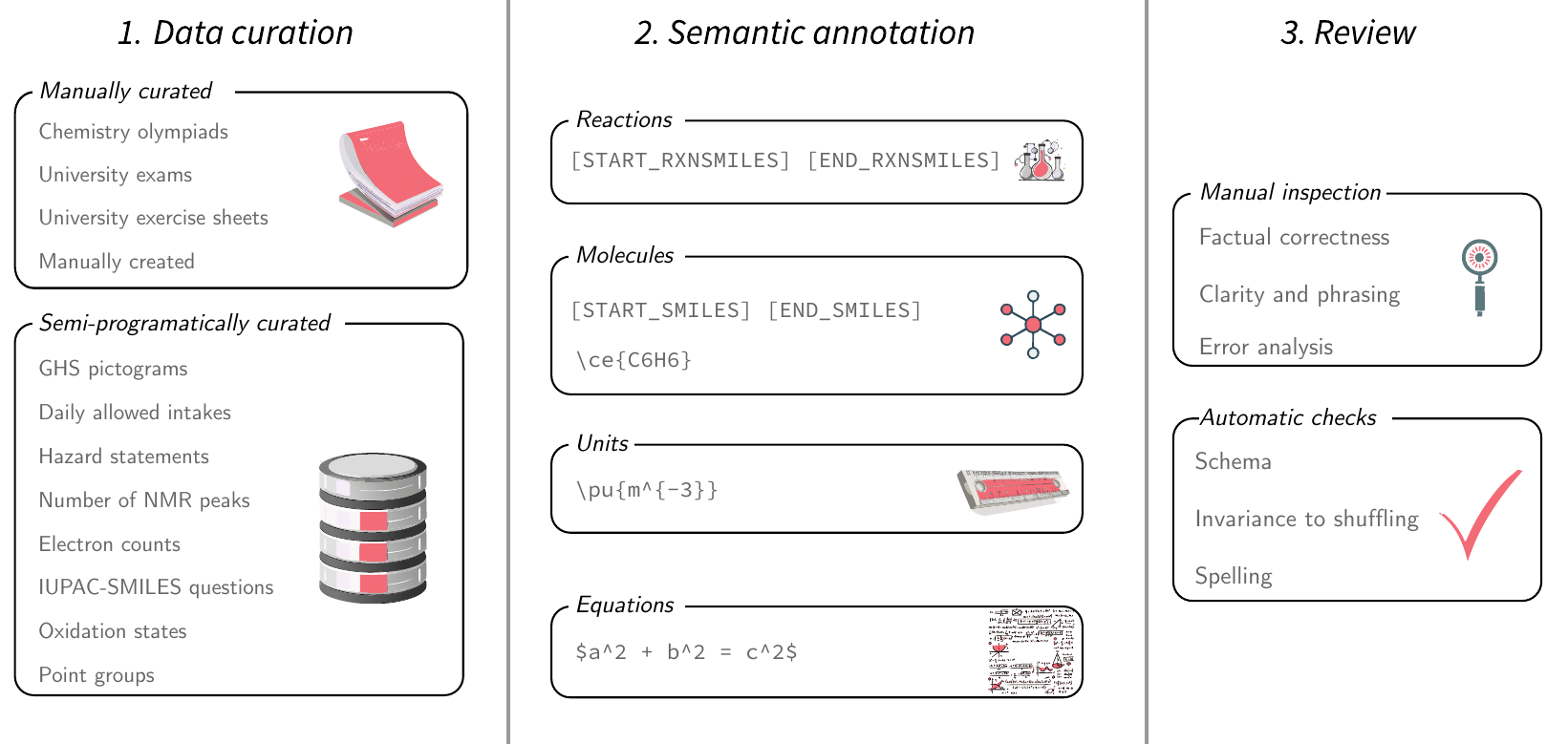}
    \caption{\textbf{Overview of the workflow for the assembly of the \chembench corpus}.
    To assemble the \chembench corpus, we first collected questions from various sources. Some tasks were manually curated, others semi-programmatically. We added semantic annotations for all questions to make them compatible with systems that use special processing for modalities that are not conventional natural text. We reviewed the questions using manual and automatic methods before adding them to the corpus.}
    \label{fig:curation_workflow}
\end{figure}

For our dataset, we curated questions from existing exams or exercise sheets but also programmatically created new questions (\Cref{tab:sources}).
Questions were added via Pull Requests on our GitHub repository and only merged into the corpus after passing manual review (\Cref{fig:curation_workflow}) as well as automated checks (e.g., for compliance with a standardized schema).

To ensure that the questions do not enter a training dataset, we use the same canary string as the BigBench project.
This requires that \Gls{llm} developers filter their training dataset for this canary string.\autocite{openai2024gpt4, srivastava2022beyond}

\begin{table}[!h]
    \centering
    \caption{\textbf{Overview of sources of the curated questions}. The table provides an overview of the types of sources the questions have been curated from. Detailed sources are available in the source data on GitHub. Questions without a source have been curated completely from scratch. Questions based on lecture notes or URLs have been curated based on content presented in those resources. All questions have been rephrased, annotated, and reviewed before being added to the corpus. }
    %
  \begin{tabular}{lc}
\toprule
Source & Count \\
\midrule
Semiautomatically generated & 1749 \\
URL & 375 \\
Textbook & 206 \\
Exam & 149 \\
IChO & 149 \\
No source & 139 \\
Lectures & 21 \\
\bottomrule
\end{tabular}
\unskip\label{output/sources_table.tex}\unskip%

    \label{tab:sources}
\end{table}

\paragraph{Manually curated questions}
Manually curated questions were sourced from various sources, including university exams, exercises, and question banks.

\paragraph{Semi-programmatically generated questions}
In addition to the manually curated questions, we also generated questions programmatically. An overview of the sources of the semi-programmatically generated questions is provided in \Cref{tab:semi_programatically_sources}.

\begin{xltabular}{\textwidth}{p{3.5 cm}p{6.5 cm}p{.5cm}X}
    \caption{\textbf{Sources of semi-programatically generated questions.} The table shows the sources and a brief description as well as the number of the semi-programatically generated questions.}\label{tab:semi_programatically_sources} \\

    \toprule
    source & description & question count \\
\midrule
Number of isomers & MAYGEN\autocite{Yirik_2021} was used to compute the number of isomers for a set of \gls{smiles} extracted from the ZINC dataset\autocite{Irwin_2012} & & %
  24 \unskip\label{output/question_count_per_dir/json_file_counts_number_of_isomers.txt}\unskip%
 \\
\midrule
Total electron count of molecules & Electron counts based on the data from  \url{https://www.cheminfo.org/} & &%
  25 \unskip\label{output/question_count_per_dir/json_file_counts_electron_counts.txt}\unskip%
 \\
\midrule
Oxidation states & Oxidation states questions based on the data from \url{https://www.cheminfo.org/} && %
  10 \unskip\label{output/question_count_per_dir/json_file_counts_oxidation_states.txt}\unskip%
 \\
\midrule
Chemical reactivity & Questions are framed based on the information from the \href{https://cameochemicals.noaa.gov/reactivity}{Cameo Chemicals website} && %
  276 \unskip\label{output/question_count_per_dir/json_file_counts_reactive_groups.txt}\unskip%
 \\
\midrule
Number of \gls{nmr} signals & Molecules are sampled from the ZINC database\autocite{Irwin_2012}, OpenChemLib\autocite{openchemlib} is used to compute the number of diasterotopically distinct hydrogen atoms && %
  50 \unskip\label{output/question_count_per_dir/json_file_counts_number_of_nmr_peaks.txt}\unskip%
 \\
\midrule
Point group of molecules & Our \href{https://github.com/lamalab-org/chem-caption}{ChemCaption} tool is used to assign the point group using spglib,\autocite{spglib}  and then each case was manually checked to select well-defined cases && %
  16 \unskip\label{output/question_count_per_dir/json_file_counts_point_group.txt}\unskip%
 \\
\midrule
IUPAC-SMILES pairs & Sampled from the PubChem \autocite{pubchem}  database && %
  10 \unskip\label{output/question_count_per_dir/json_file_counts_smiles_to_name.txt}\unskip%
 + %
  10 \unskip\label{output/question_count_per_dir/json_file_counts_smiles_to_name.txt}\unskip%
 \\
\midrule
\multirow{3}{*}{PubChem \autocite{pubchem}  safety data} & Daily allowable intakes according to the World Health Organization && %
  10 \unskip\label{output/question_count_per_dir/json_file_counts_dai.txt}\unskip%
  \\
 & Definitions of hazard statements &&  %
  10 \unskip\label{output/question_count_per_dir/json_file_counts_h_statements.txt}\unskip%
 \\
 & GHS classification of chemicals mined through the API& & %
  7 \unskip\label{output/question_count_per_dir/json_file_counts_pictograms.txt}\unskip%
 \\
 \midrule
\multirow{2}{*}{Safety}
& Materials' compatibility && %
  20 \unskip\label{output/question_count_per_dir/json_file_counts_materials_compatibility.txt}\unskip%
 \\
 & Chemical compatibility && %
  296 \unskip\label{output/question_count_per_dir/json_file_counts_chem_chem_comp.txt}\unskip%
 \\
\bottomrule
\end{xltabular}

\paragraph{Chemical preference data}

These questions assess the ability to establish a \enquote{preference}, such as favoring a specific molecule. Chemical preference is of major importance in drug discovery projects, where the optimization process to reach the desired molecular properties is a process that takes several years within a chemist's career.
Our data corpus is adapted from the published dataset by \textcite{Choung_2023}, which consists of more than 5000 question-answer pairs about chemical intuition. To build the dataset, they presented 35 medicinal chemists with two different molecules, asking them what molecule they would like to continue with when imaging an early virtual screening campaign setting. The question was designed so the scientists do not spend much time answering it, relying only on their feelings or \enquote{chemical preference}.

To understand whether the capabilities of the leading models align with the preferences of professional chemists, we randomly selected 1000 data points from the original dataset to create a meaningful evaluation set, where molecules are represented as \gls{smiles}.
To ablate the effect of different molecular representations, we only considered questions for which we could obtain \gls{iupac} names for both molecules present.

\subsection{Model evaluation workflow}
A graphical overview of the pipeline is shown in \Cref{fig:process}.

\paragraph{Prompting}

We employ distinct prompt templates tailored for completion and instruction-tuned models to maintain consistency with the training.
As explained below, we impose constraints on the models within these templates to receive responses in a specific format so that robust, fair, and consistent parsing can be performed.
Certain models are trained with special annotations and \LaTeX\xspace syntax for scientific notations, chemical reactions, or symbols embedded within the text.
For example, all the \gls{smiles} representations are encapsulated within \texttt{[START\_SMILES][\textbackslash END\_SMILES]} in Galactica\autocite{taylor2022galactica}.
Our prompting strategy consistently adheres to these details in a model-specific manner by post-processing \LaTeX\xspace syntax, chemical symbols, chemical equations, and physical units (by either adding or removing wrappers).
This step can be easily customized in our codebase, and we provide presets for the models we evaluated.

\paragraph{Parsing}
Our parsing workflow is multistep and primarily based on regular expressions.
In the case of instruction-tuned models, we first identify the \texttt{[ANSWER]}\texttt{[\textbackslash ANSWER]} environment we prompt the model to report the answer in.
In the case of completion models, this step is skipped. From there, we attempt to extract the relevant enumeration letters (for multiple-choice questions) or numbers.
In the case of numbers, our regular expression was engineered to deal with various forms of scientific notation.
As initial tests indicated that models sometimes return integers in the form of words, e.g., \enquote{one} instead of \enquote{1}, we also implemented a word-to-number conversion using regular expressions.
If these hard-coded parsing steps fail, we use a \gls{llm}, e.g., \ClaudeThreeFiveSonnet, to parse the completion (\Cref{sec:llm-parsing} provides more details on this step).

\paragraph{Models}
For all models, we performed inference using greedy decoding (i.e., temperature 0). We used the \gls{api} endpoints provided by the model developers and those provided by Groq. \PaperQATwo was used (in August 2024) via an \gls{api} provided by FutureHouse.

\subsection{Confidence estimate}
To estimate the models' confidence, we prompted them with the question (and answer options for \gls{mcq}) and the task to rate their confidence to produce the correct answer on a scale from 1 to 5.
We decided to use verbalized confidence estimates\autocite{xiong2023llms} since we found those closer to current practical use cases than other prompting strategies, which might be more suitable when implemented in systems. In addition, this approach captures semantic uncertainty, which is not the same as the probability of a token being given a sequence of tokens (i.e., the uncertainty one obtains from logit-based approaches). On top of that, many proprietary models do not provide access to the logits, making this approach more general.
In \Cref{sec:confidence_estimates}, we provide more details and comparisons with a logit-based approach.

\subsection{Human baseline}

\paragraph{Question selection} \label{sec:subset-selection}

 Several design choices were made when selecting \chembenchmini. Firstly, from the full dataset, we kept all the questions labeled as advanced. In this way, we can obtain a deeper insight into the capabilities of \glspl{llm} on advanced tasks when compared to actual chemists. Secondly, we sample a maximum of three questions across all possible combinations of categories (i.e., knowledge or reasoning) and topics (e.g., organic chemistry, physical chemistry). Thirdly, we do not include any intuition questions in this subset because the intended use of \chembenchmini is to provide a fast and fair evaluation of \glspl{llm} independent of any human baseline. In total, %
  236 \unskip\label{output/num_human_answered_questions.txt}\unskip%
 questions have been sampled for \chembenchmini. Then, this set is divided into two subsets based on the aforementioned combinations. One of the question subsets allows tool use, and the other does not.

\paragraph{Study design}
Human volunteers were asked the questions in a custom-built web interface (see \Cref{sec:human_baseline}), which rendered chemicals and equations. Questions were shown in random order, and volunteers were not allowed to skip questions. For a subset of the questions, the volunteers were allowed to use external tools (excluding other \gls{llm} or asking other people) to answer the questions. Prior to answering questions, volunteers were asked to provide information about their education and experience in chemistry. The study was conducted in English.

\paragraph{Human volunteers}
Users were open to reporting about their experience in chemistry.
Overall, %
  16 \unskip\label{output/num_users_with_education_info.txt}\unskip%
 did so.
Out of those, %
  2 \unskip\label{output/num_human_postdoc.txt}\unskip%
 are beyond a first postdoc, %
  13 \unskip\label{output/num_human_master.txt}\unskip%
 have a master's degree (and are currently enrolled in Ph.D.\ studies), and %
  1 \unskip\label{output/num_human_bachelor.txt}\unskip%
 has a bachelor's degree. For the analysis, we excluded volunteers with less than two years of experience in chemistry after their first university-level course in chemistry.

\paragraph{Comparison with models}
For the analysis, we treated each human as a model. We computed the topic aggregated averages per human for analyses grouped by topic and then averaged over all humans. The performance metrics reported for models in the main text are computed on the same questions the humans answered. Metrics for the entire corpus are reported in the appendix (\Cref{sec:model_performance_app}).

\subsection{Data annotation}\label{sec:meth-topic}
In the curation of our dataset, we manually assigned difficulty levels and required skills to each question. We used the following guidelines for these annotations: calculation is required if answering a question would require the use of a calculator, knowledge is required if answering a question requires non-trivial knowledge of facts (e.g., the H/P statements of chemicals). Reasoning is required if answering a question requires multiple reasoning steps.
Basic questions only require those skills up to the high school level. Advanced questions would require an expert multiple minutes up to hours to answer.

\section*{Data and code availability}
The code and data for \chembench are available at \url{https://github.com/lamalab-org/chem-bench} and archived on Zenodo under \href{10.5281/zenodo.14010212}{https://zenodo.org/records/14010212}.
The code for the app for our human baseline study is available at \url{https://github.com/lamalab-org/chem-bench-app}.
To ensure reproducibility, this manuscript was generated using the \href{https://show-your.work/en/latest/}{\showyourwork} framework.\autocite{Luger2021}
The code to rebuild the paper (including code for all figures and numbers next to which there is a GitHub icon) can be found at \url{\GitHubURL}.
To facilitate reproduction, some intermediate analysis results are cached at \url{http://dx.doi.org/10.5072/zenodo.34706}.

\section*{Acknowledgements}
This work was supported by the Carl Zeiss Foundation, and a \enquote{Talent Fund} of the \enquote{Life} profile line of the Friedrich Schiller University Jena.

In addition, M.S-W.'s work was supported by Intel and Merck via the AWASES programme.

Parts of A.M.'s work was supported as part of the \enquote{SOL-AI} project funded by the Helmholtz Foundation model initative.

K.M.J.\ is part of the NFDI consortium FAIRmat funded by the Deutsche Forschungsgemeinschaft (DFG, German Research Foundation) – project 460197019.

K.M.J.\ thanks FutureHouse (a non-profit research organization supported by the generosity of Eric and Wendy Schmidt) for supporting \PaperQATwo runs via access to the \gls{api}. We also thank Stability.AI for the access to its HPC cluster.

M.R.G.\ and M.V.G.\ acknowledge financial support from the Spanish Agencia Estatal de Investigaci\'{o}n (AEI) through grants TED2021-131693B-I00 and CNS2022-135474, funded by MICIU/AEI/10.13039/501100011033 and by the European Union NextGenerationEU/PRTR. M.V.G.\ acknowledges support from the Spanish National Research Council (CSIC) through Programme for internationalization i-LINK 2023 (Project ILINK23047).

A.A.\ gratefully acknowledges financial support for this research by the Fulbright U.S. Student Program, which is sponsored by the U.S.\ Department of State and German-American Fulbright Commission. Its contents are solely the responsibility of the author and do not necessarily represent the official views of the Fulbright Program, the Government of the United States, or the German-American Fulbright Commission.

M.A.\ expresses gratitude to the European Research Council (ERC) for evaluating the project with the reference number 101106377 titled \enquote{CLARIFIER} and accepting it for funding under the HORIZON TMA MSCA Postdoctoral Fellowships - European Fellowships.
Furthermore, M.A.\ acknowledges the funding provided by UK Research and Innovation (UKRI) under the UK government’s Horizon Europe funding guarantee (Grant Reference: EP/Y023447/1; Organization Reference: 101106377).

M.R.\ and U.S.S.\ thank the \enquote{Deutsche Forschungsgemeinschaft} for funding under the regime of the priority programme SPP 2363 \enquote{Utilization and Development of Machine Learning for Molecular Applications – Molecular Machine Learning} (SCHU 1229/63-1; project number 497115849).

In addition, we thank the OpenBioML.org community and their ChemNLP project team for valuable discussions.
Moreover, we thank Pepe Márquez for discussions and support and Julian Kimmig for feedback on the web app.
In addition, we acknowledge support from Sandeep Kumar with an initial prototype of the web app.
We thank Bastian Rieck for developing the \LaTeX-credit package (\url{https://github.com/Pseudomanifold/latex-credits}) and thank Berend Smit for feedback on an early version of the manuscript.

\section*{Statement of ethical compliance}
The authors confirm to have complied with all relevant ethical regulations, according to the Ethics Commission of the Friedrich Schiller University Jena (which decided that study is ethically safe). Informed consent was obtained from all volunteers.

\section*{Conflicts of interest}
K.M.J.\ was a paid consultant for OpenAI (as part of the red teaming network). M.P.\ is an employee of Stability.AI, and A.M.\ and N.A.\ were paid contractors of Stability.AI.

\section*{Author contributions}

\resizebox{\textwidth}{!}{%
\scriptsize
\insertcredits
}
\normalsize
\printbibliography
\end{refsection}

\clearpage
\begin{refsection}
\renewcommand{\thefigure}{A\arabic{figure}}
\setcounter{figure}{0}

\renewcommand{\thetable}{A\arabic{table}}
\setcounter{table}{0}

\appendix
\section{Appendix}

\subsection{Desired properties of a chemistry benchmark} \label{sec:desired-properties}

\begin{itemize}
    \item \emph{End-to-end automation}. For model development, the evaluations must be run many times (e.g., on regular intervals of a training run).
    Approaches that rely on humans scoring the answers of a system\autocite{Schulze_Balhorn_2024, ai4science2023impact, castro2023large} can thus not be used.
    \item \emph{Careful validation by experts}. Manual curation is needed to minimize the number of incorrect or unanswerable questions.\autocite{northcutt2021pervasive}
    This is motivated by the observation that many widely used benchmarks are plagued by noisiness.\autocite{Frye_2023, Awg}
    \item \emph{Usable with models that support special treatment of molecules}. Some models, such as Galactica\autocite{taylor2022galactica}, use special tokenization or encoding procedures for molecules or equations.
    The benchmark system must encode the semantic meaning of various parts of the question or answer to support this.
    \item \emph{Usable with black box systems}. Many relevant systems do not provide access to model weights or raw logits.
    This might be the case because the systems are proprietary or because they involve not only \glspl{llm} but also external tools such as search \glspl{api} or code executors.\autocite{schick2024toolformer, karpas2022mrkl, yao2022react}
    Thus, a benchmark should not assume access to the raw model outputs but be able to operate on text completions.
    \item \emph{Probing capabilities beyond answering of \glspl{mcq}}. In real-world chemistry, as well as higher-level university education, multiple-choice questions are seldom utilized.
    Yet, most benchmarking frameworks focus on the \gls{mcq} setting because of the ease of evaluation. Realistic evaluations must measure capabilities beyond answering \gls{mcq}.
    \item \emph{Cover a diverse set of topics}. Chemistry, as the \enquote{central science}, bridges multiple disciplines.\autocite{Aspuru_Guzik_2018} To even just approximate \enquote{chemistry capabilities}, the topics covered by a chemistry benchmark must be very diverse.
    \item \emph{Cover diverse skills}. To holistically judge performance, it is important to cover questions that rely on reasoning, calculation, knowledge, and intuition.
    \item \emph{Cover a range of difficulty levels}. To allow for a continuous measure of improvement for a range of different (evolving) systems, a benchmark should cover a wide range of difficulty levels.
    \item \emph{Impossible to completely solve with current models}. A benchmark should contain questions that are impossible to solve with current models. The benchmark provides no useful signal if current models can solve all questions.
\end{itemize}

\subsection{Related work}
Existing benchmarks such as those from \textcite{guo2023large}, \textcite{sun2023scieval}, \textcite{Schulze_Balhorn_2024}, \textcite{Cai_2024}, \textcite{rein2023gpqagraduatelevelgoogleproofqa} fail to comply with most of the requirements stipulated above.
While these benchmarks could provide valuable insights in the short term, they cannot follow the rapid additions to the \gls{llm} space.
\chembench aims to correct this through a set of developments: compatibility with BigBench, end-to-end automation, a particular focus on chemical safety, employment of diverse prompting strategies, and specialized notation for molecules and mathematical symbols.
Moreover, our robust framework, including the platform \url{chembench.org}, will engage the community in open-source contributions.

\clearpage
\subsection{Benchmark corpus}
To ensure maximal interoperability with existing benchmarks or tools, we curated the data in an extended form of the widely used BigBench format.\autocite{srivastava2022beyond}
This also implies that future baselines can be built on top of our infrastructure if saved in the same format.

\subsubsection{Curation workflow}
Questions were added via pull requests to the \chembench repository on GitHub.
This allowed for a manual review of each question by expert reviewers (with backgrounds in chemistry, materials science, chemical engineering, and physics).
The reviews were conducted directly on the GitHub platform, where our entire curation history is also publicly available.

The general guidelines followed by the reviewer are the following:

\begin{itemize}
    \item \textit{Originality:} Questions should not be readily findable online or in other easily accessible sources (example \url{https://github.com/lamalab-org/chem-bench/pull/392#discussion_r1694299474})
    \item \textit{Ambiguity:} Questions with unclear wording or multiple interpretations (example \url{https://github.com/lamalab-org/chem-bench/pull/420#discussion_r1698147159} )
    \item \textit{Factual or heuristic Errors: }Questions containing factual inaccuracies or misconceptions are not included (example \url{https://github.com/lamalab-org/chem-bench/pull/389#discussion_r1686187301})
    \item \textit{Clarity and Difficulty: } They should pose a challenge and encourage exploration within the chemical domain (example \url{https://github.com/lamalab-org/chem-bench/pull/391#discussion_r1679276714})
    \item \textit{Out of Scope:} Questions outside the realm of chemistry are rejected.
    \item \textit{Contribution to Dataset Diversity: }Questions should cover a wide range of chemical concepts and sub-disciplines. They should add value by expanding the breadth of the dataset. That is, questions already multiple (>10) times in the corpus in a similar form are rejected.
\end{itemize}

Reviewers also solved the questions to verify the answers. They also performed web searches to ensure questions were not easily found online. The reviewers often guided the revision process to ensure the question aligned with the guidelines. Questions that don't meet the criteria are either rejected or suggested for revision, and most often, they are modified to a new question. Reviewers also provide feedback on the skill and difficulty annotations.

In addition to the manual review, we also performed automated checks to ensure the quality of the questions. The schemas, \LaTeX\xspace templating, and other formatting aspects are checked automatically using GitHub Actions.

While adding questions from existing benchmarks might seem to be another good source of semi-automatically generated data, we prioritized the diversity of the data and avoided data contamination while addressing the guidelines above and in \Cref{sec:desired-properties}.
However, even though we decided not to include questions from other previously published chemistry-focused benchmarks into the \chembench corpus, the framework is flexible enough to be readily extended with questions from other benchmarks.

\subsubsection{Composition}

\begin{figure}
    \centering
    \includegraphics[width=\textwidth]{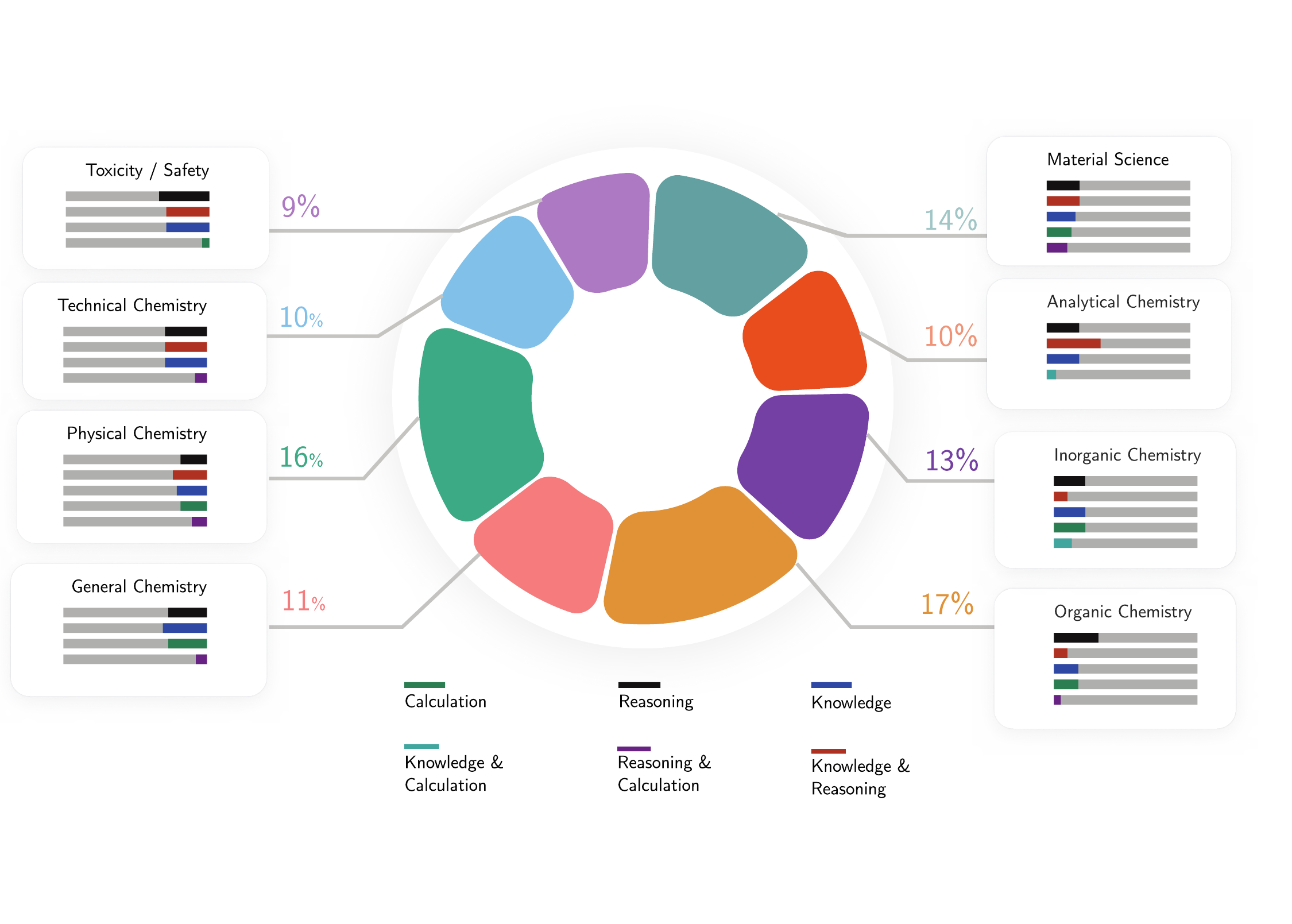}
    \caption{\textbf{Composition of the human subset.} The circular plot shows the distribution of topics and required skills in the human subset of the \chembench corpus. The human subset is a representative subset of the full corpus, with a balanced distribution of topics and skills.}
    \label{fig:cb_humanset}
\end{figure}

\Cref{fig:cb_humanset} shows the distribution of topics and required skills in the human subset of the \chembench corpus.

The corpus of the questions in \chembench, as shown in \Cref{tab:chembench_corpus_topic}, can be divided according to which chemical topic they belong.

\begin{xltabular}{\textwidth}{X}
    \caption{\textbf{Examples for each of the topics considered in the evaluation of the \chembench corpus.} The table shows the percentage of questions in the corpus that belong to each topic, as well as example questions.} \label{tab:chembench_corpus_topic} \\
            \toprule
            \multicolumn{1}{c}{\textbf{Analytical Chemistry} %
  148 Questions (38.51\%) \unskip\label{output/total_analytical.txt}\unskip%
} \\
            \midrule
            Which of the following analytical methods is most appropriate for performing a survey analysis of a solid sample containing various metals? \\
            A. X-ray fluorescence analysis \\
            B. Differential pulse polarography \\
            C. Flame-atomic absorption spectroscopy \\
            D. Gas chromatography with flame ionization detector \\
            E. Hydride generation atomic absorption spectroscopy \\
            \midrule
            \multicolumn{1}{c}{\textbf{Chemical Preference} %
  1001 Questions (51.15\%) \unskip\label{output/total_chemical_preference.txt}\unskip%
} \\
            \midrule
            Imagine an early virtual screening campaign setting (accounting for simple aspects such as oral availability and small molecular profile, but no other modalities such as covalency or bifunctionality). Which of the following two candidates would you prefer for further development? \\
            A. [START\_SMILES]N\#Cc1ccc(OCCCN2CC3CN\-(CCNS(=O)(=O)c4ccc(F)cc4)CC(C2)O3)cc1\-[END\-\_SMILES] \\
            B. [START\_SMILES]O=C1CC(c2ccc(CC(NS(=O)\-(=O)\-c3cc(Cl)cc(Cl)c3)c3nc4ccccc4[nH]3)cc2)S\-(=O)\-(=O)N1[END\_SMILES] \\
            \midrule
            \multicolumn{1}{c}{\textbf{General Chemistry} %
  150 Questions (50.67\%) \unskip\label{output/total_general.txt}\unskip%
} \\
            \midrule
            Which of the following salts is an acidic salt? \\
            A. \ce{NH4Cl} \\
            B. \ce{Na2CO3} \\
            C. \ce{NaH2PO4} \\
            D. \ce{Zn(OH)Cl} \\
            \midrule
            \multicolumn{1}{c}{\textbf{Inorganic Chemistry} %
  101 Questions (56.44\%) \unskip\label{output/total_inorganic.txt}\unskip%
} \\
            \midrule
            What is the oxidation number of the metal in the compound \ce{[ZrF_{7}]3-}? \\
            \midrule
            \multicolumn{1}{c}{\textbf{Materials Science} %
  10 Questions (30.00\%) \unskip\label{output/total_materials_science.txt}\unskip%
} \\
            \midrule
            For NMR analysis, you need to digest the MOF in a strong acid to remove the linker and leave the metal clusters intact. Why would one choose HF over \ce{HCl} for this purpose? \\
            A. \ce{F-} forms a stable bonds to the metal ions \\
            B. \ce{HF} has a better water solubility than \ce{HCl} \\
            C. \ce{HF} has a higher boiling point than \ce{HCl} \\
            D. \ce{HF} is a weaker acid than \ce{HCl} \\
            \midrule
            \multicolumn{1}{c}{\textbf{Organic Chemistry} %
  334 Questions (59.58\%) \unskip\label{output/total_organic.txt}\unskip%
} \\
            \midrule
            What is the reaction mechanism that describes the following reaction (represented using reaction SMILES) [START\_RXNSMILES]CCCl.CO[Na]>>[Na]Cl.CCOC\-[END\_RXNSMILES]? \\
            A. $E_1$ \\
            B. $E_{cb}$ \\
            C. $S_N1$ \\
            D. $S_N2$ \\
            \midrule
            \multicolumn{1}{c}{\textbf{Physical Chemistry} %
  142 Questions (53.52\%) \unskip\label{output/total_physical.txt}\unskip%
} \\
            \midrule
            The Born-Oppenheimer (BO) approximation is widely used in computational chemistry, but its accuracy can vary depending on the system. Among the following options, for which system is the Born-Oppenheimer approximation likely to be least applicable? \\
            A. \ce{C60} \\
            B. \ce{CH4} \\
            C. \ce{Fe(CO)5} \\
            D. \ce{H2+} \\
            E. \ce{NaCl} \\
            \midrule
            \multicolumn{1}{c}{\textbf{Technical Chemistry} %
  40 Questions (47.50\%) \unskip\label{output/total_technical.txt}\unskip%
} \\
            \midrule
            Which of the following statements is true about the different types of ideal reactors? \\
            A. In a batch reactor, the composition is uniformly mixed and remains the same throughout the reactor and at the exit \\
            B. In a batch reactor, the fluid passes through the reactor with no mixing of earlier and later entering fluid \\
            C. In a mixed flow reactor, the composition changes with time but is uniform everywhere within the reactor \\
            D. In a plug flow reactor, the fluid moves in single flow through the reactor with no mixing and no overtaking \\
            \midrule
            \multicolumn{1}{c}{\textbf{Toxicity/Safety} %
  675 Questions (33.19\%) \unskip\label{output/total_toxicity_safety.txt}\unskip%
} \\
            \midrule
            Pindolol and propranolol are (relatively nonselective) antagonists at $\beta_1$- and $\beta_2$-adrenoceptors. However, pindolol is a partial agonist, whereas propranolol is a pure antagonist. What follows from this? \\
            A. Pindolol has a greater therapeutic range than propranolol \\
            B. Pindolol has a longer half-life than propranolol \\
            C. Pindolol has intrinsic activity \\
            D. Pindolol is more lipophilic than propranolol \\
            E. Pindolol is more potent than propranolol \\
            \bottomrule
 \end{xltabular}

\normalsize

\clearpage

\begin{figure}
    \centering
    \includegraphics[width=\textwidth]{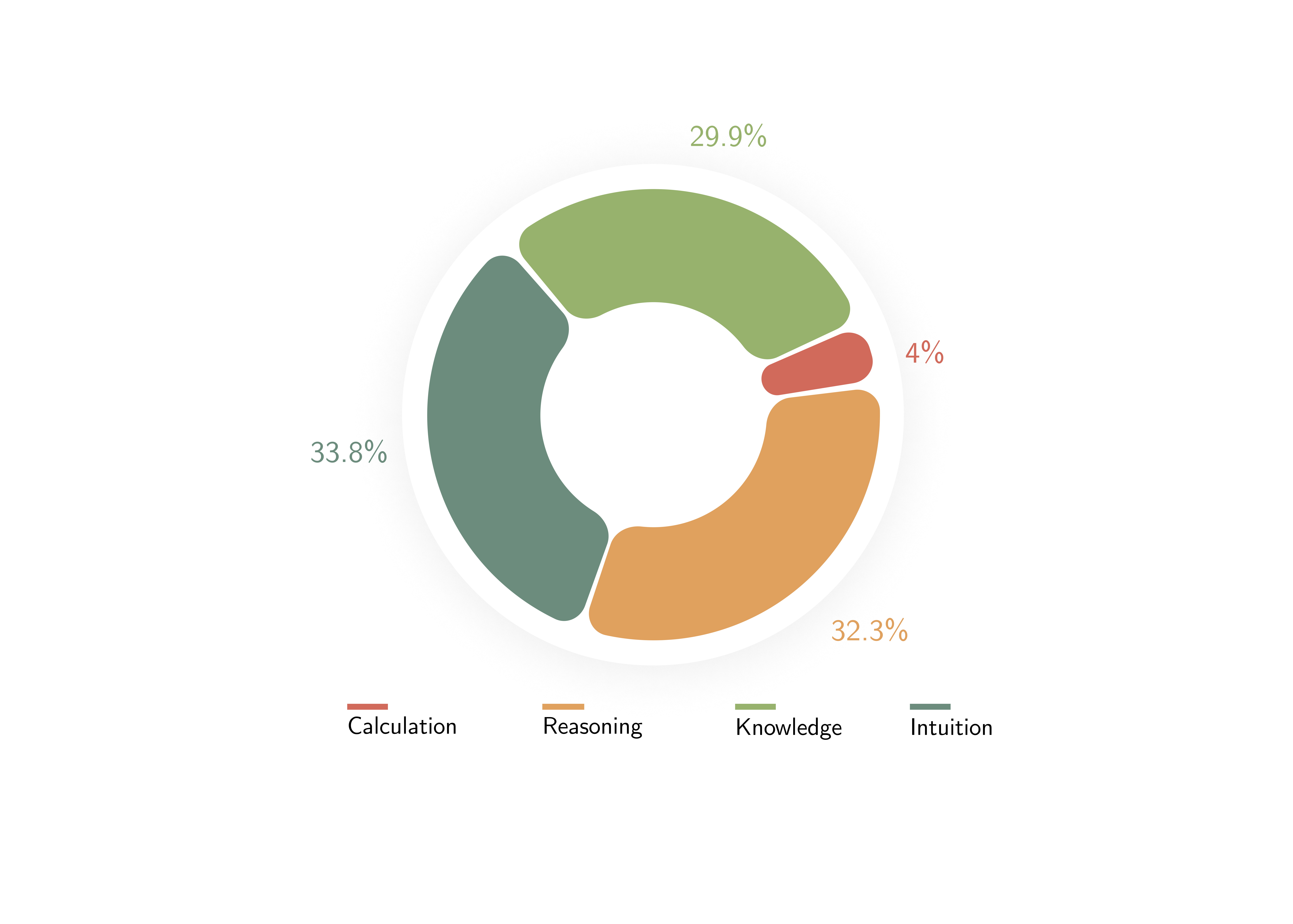}
    \caption{\textbf{Composition of the required skills considered in the \chembench corpus.} The circular plot shows the distribution of required skills in the \chembench corpus.}
    \label{fig:cb_skillset}
\end{figure}

In addition, as shown in \Cref{fig:cb_skillset} the \chembench corpus can be divided considering the different skills needed to solve the questions. The plot shows a balanced distribution of the required skills in the \chembench corpus. \Cref{tab:chembench_corpus_cognitive} shows an example question for each skill category.

\begin{xltabular}{\textwidth}{X}
    \caption{\textbf{Examples for each required skill considered in the \chembench corpus.} The table shows the number of questions for each skill and an example question. Note that the total count in this table is bigger than the \chembench corpus. This is because the same question can be annotated with two different skills, e.g., Reasoning and Calculation.}    \label{tab:chembench_corpus_cognitive} \\
            \toprule
            \multicolumn{1}{c}{\textbf{Knowledge} %
  886 \unskip\label{output/knowledge_count.txt}\unskip%
\hspace{1.5em} Questions} \\
            \midrule
            Which of the following salts is an acidic salt? \\
            A. \ce{NH4Cl} \\
            B. \ce{Na2CO3} \\
            C. \ce{NaH2PO4} \\
            D. \ce{Zn(OH)Cl} \\
            \midrule
            \multicolumn{1}{c}{\textbf{Reasoning} %
  955 \unskip\label{output/reasoning_count.txt}\unskip%
\hspace{1.5em} Questions} \\
            \midrule
            You make the following experimental observations: The sample consisted of small colorless/white crystals, wine-red transparent crystals, turquoise transparent crystals, and violet flakes. Most components dissolved in cold water, with turquoise crystals dissolving upon heating. Dimethylglyoxime produced a raspberry-red color with dissolved turquoise crystals. Violet flakes turned blue on contact and showed red/violet colors in flame tests. An oxidation melt turned green. No red coloration occurred with \ce{KSCN}, but a blue organic phase formed with acetone. Potassium ferrocyanide showed no changes. \ce{KOH} produced a brown/black precipitate. Alizarin-S testing resulted in red coloration after acidification. Heating with cobalt nitrate produced a bluish color without luster. Lead dioxide and sulfuric acid oxidation formed a violet solution. No changes occurred in chromate oxidation or titanium detection attempts. \ce{BaCl2} formed precipitates in both original solution and soda extract. Lunge's reagent showed red coloration around zinc granules. Silver nitrate formed white precipitates that dissolved in \ce{NH3} and reappeared with \ce{HNO3} acidification. The separation procedure began with the aqueous solution being treated with \ce{(NH4)2CO3} solution until a persistent turbidity appeared, followed by addition of dilute \ce{HCl} until the turbidity disappeared. The pH was then adjusted to 5-6 with \ce{NH3} (2 M) and heated on a water bath. Three spatula tips of urotropin were added, resulting in a white, rather flocculent precipitate. This precipitate was centrifuged off and washed with warm water. In the Urotropin group, the residue was dissolved in dilute \ce{HCl} and diluted with water. An alkaline precipitation was performed, resulting in a brown coloration even after repeated washing of the urotropin precipitation. This precipitate dissolved in 2 M \ce{HCl}, but showed no change when treated with either \ce{KSCN} or potassium ferrocyanide. In the \ce{(NH4)2S} group, addition of dimethylglyoxim produced a very flocculent red precipitate that rose to the top of the test tube. When \ce{H2S} was introduced to the centrifugate, a black precipitation occurred. The precipitates did not completely dissolve in \ce{HCl}. Treatment with \ce{KSCN} resulted in a blue organic phase when layered with acetone. An alkaline precipitation produced a brown/black residue. No precipitation was observed when \ce{H2S} was introduced to the supernatant liquid after the alkaline precipitation.  What ions are in the sample? \\
            A. \ce{Al^{3+}} \\
            B. \ce{Co^{2+}} \\
            C. \ce{Cr^{3+}} \\
            D. \ce{Fe^{3+}} \\
            E. \ce{Mn^{2+}} \\
            F. \ce{Ni^{2+}} \\
            G. \ce{Zn^{2+}} \\
            \midrule
            \multicolumn{1}{c}{\textbf{Intuition} %
  1001 \unskip\label{output/intuition_count.txt}\unskip%
\hspace{1.5em} Questions} \\
            \midrule
            Imagine an early virtual screening campaign setting (accounting for simple aspects such as oral availability and small molecular profile, but no other modalities such as covalency or bifunctionality). Which of the following two candidates would you prefer for further development? \\
            A. [START\_SMILES]CC1(C)Oc2ccc([N+](=O)[O-])cc2[C@@H](N2CCOCC2)[C\@\@H]1O[END\_SMILES] \\
            B. [START\_SMILES]Cc1ccccc1N=C(S)N1CCC(NC\-(=O)c2ccco2)CC1[END\_SMILES] \\
            \midrule
            \multicolumn{1}{c}{\textbf{Calculation} %
  118 \unskip\label{output/calculation_count.txt}\unskip%
\hspace{1.5em} Questions} \\
            \midrule
            Given that the average molar mass of the polymer chains in this sample of poly(lactic acid) (PLA) is \SI{595}{g mol^{-1}} using end-group analysis, where \SI{0.1619}{g} of PLA was dissolved in \SI{25}{cm^3} of benzyl alcohol and titrated with \SI{0.0400}{mol dm^{-3}} \ce{KOH} solution. The volume of \ce{KOH} solution required to reach the endpoint was \SI{6.81}{cm^3}. What is the average number of monomer units in each polymer chain of this sample? \\
            \bottomrule
\end{xltabular}

\clearpage
\subsection{Model performance} \label{sec:model_performance_app}
\begin{figure}[htb]
    \centering
    \includegraphics{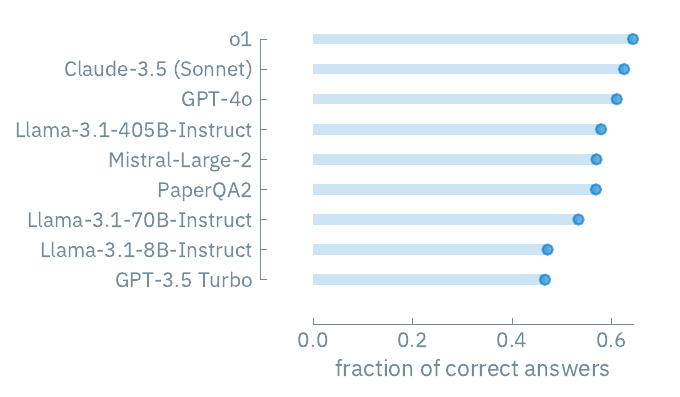}
    \caption{\textbf{Overall performance of the models on the \chembench corpus.} The bar plot shows the fraction of questions the models answered correctly. Scores computed on the entire \chembench corpus.}
    \label{fig:barplot_all_correct_all_questions}
    \script{plot_overview_performance_plot.py}
\end{figure}

We also evaluated the model performance on the entire \chembench corpus.
\Cref{fig:barplot_all_correct_all_questions} shows the fraction of questions the models answered correctly.

\begin{figure}[htb]
    \centering
    \includegraphics[width=\textwidth]{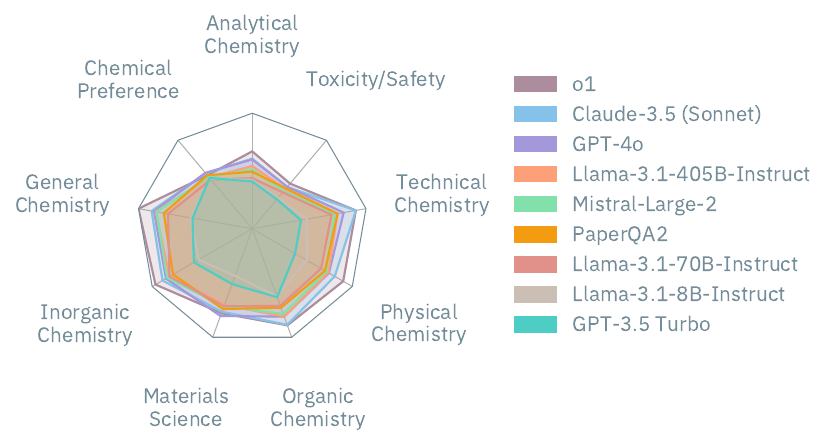}
    \caption{\textbf{Performance of the models on the different topics of the \chembench corpus.} The radar plot shows the performance of the models on the different topics of the \chembench corpus. The performance is measured as the fraction of questions answered correctly by the models.
    A score of 1 (full coverage until the outer line of this plot) indicates that all questions were answered correctly, while a score of 0 indicates that none were answered correctly.
    }
    \label{fig:all_questions_models_completely_correct_radar_overall}
    \script{analyze_model_reports.py}
\end{figure}

\Cref{fig:all_questions_models_completely_correct_radar_overall} shows the performance of the models on the different topics of the \chembench corpus.
The general pattern of performance varies significantly between the different topics and is also observed when the models are evaluated on the entire corpus.
However, since some subjects are composed of questions from different sources, the ranking of the models is, in some instances, different from the one on \chembenchmini.

\begin{figure}[htb]
    \centering
    \includegraphics{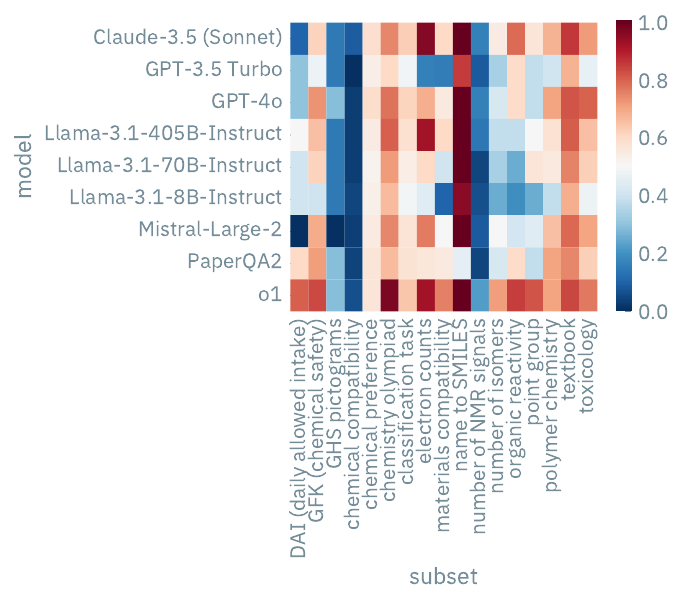}
    \caption{\textbf{Fraction of correctly answered questions per data source on \chembenchmini.} The heatmap shows, in color, the fraction of questions answered correctly by different systems for some of our data sources. The performance is measured as the fraction of questions answered correctly by the models. A score of one (red) indicates that all questions were answered correctly, while a score of zero (blue) indicates that none of the questions were answered correctly.
        We see that the performance of the models varies significantly between the different data sources. For instance, it is interesting to observe that questions sourced based on textbooks seem easier for our leading models than for humans. However, this performance does not correlate with performance on other sources, e.g., semi-programmatically created tasks such as questions about the number of signals in an \gls{nmr} spectrum.
    }
    \label{fig:performance_per_topic}
    \script{analyze_performance_per_source.py}
\end{figure}

To further investigate the performance of the models, we also compared the performance on different data sources.
Compared to topics, this is a more fine-grained analysis, as topics can be composed of questions from different sources.
In \Cref{fig:performance_per_topic}, we see that the performance of the models varies significantly between the different data sources.
Interestingly, the performance of the models on questions sourced based on textbooks seems to be better for our models than some semi-programmatically created tasks, such as questions about the number of signals in an \gls{nmr} spectrum.

\begin{figure}[htb]
    \centering
    \includegraphics{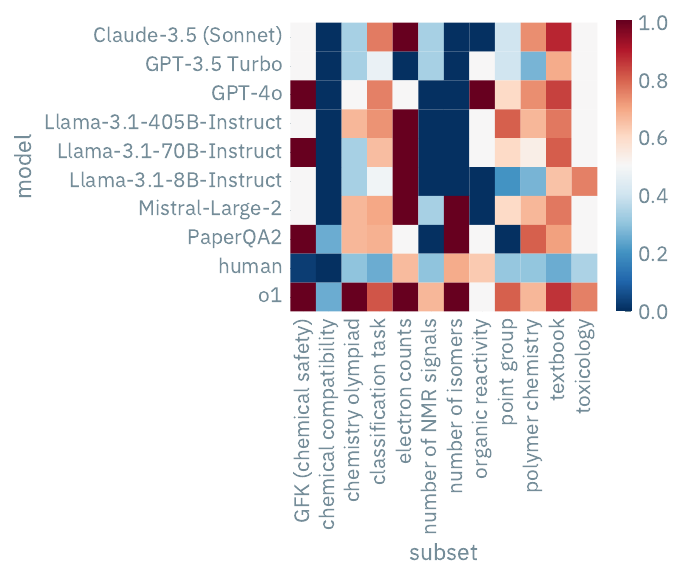}
    \caption{\textbf{Fraction of correctly answered questions per data source on \chembenchmini.} The heatmap shows, in color, the fraction of questions answered correctly by different systems for some of our data sources. The performance is measured as the fraction of questions answered correctly by the models. A score of one (red) indicates that all questions were answered correctly, while a score of zero (blue) indicates that none were answered correctly.
        We see that the performance of the models varies significantly between the different data sources. For instance, it is interesting to observe that questions sourced based on textbooks seem easier for the leading models than for humans. However, this performance does not correlate with performance on other sources, e.g., semi-programmatically created tasks such as questions about the number of signals in an \gls{nmr} spectrum.
    }
    \label{fig:performance_per_topic_tiny}
    \script{analyze_performance_per_source.py}
\end{figure}

\Cref{fig:performance_per_topic_tiny} shows the same analysis on \chembenchmini.

\begin{figure}[htb]
    \centering
    \includegraphics[width=\textwidth]{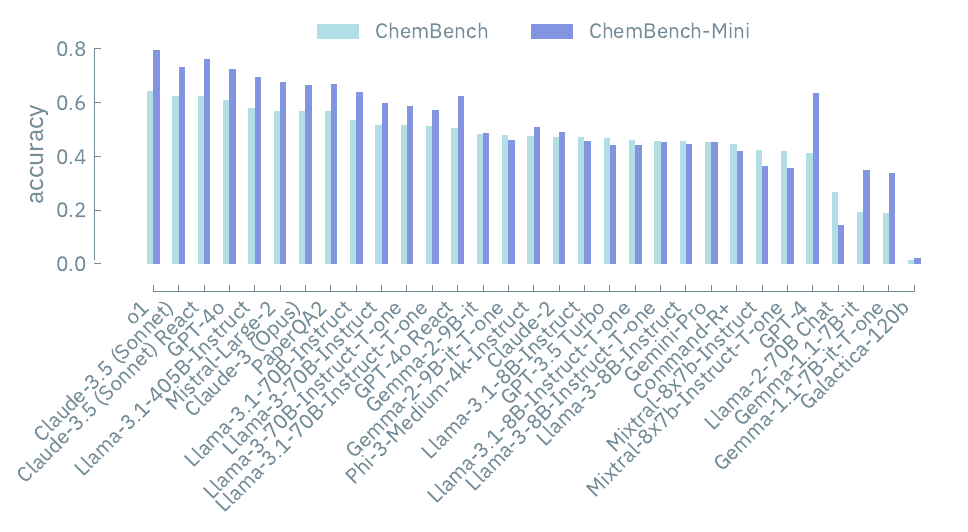}
    \caption{\textbf{
        Performance of the models on the \chembench corpus and \chembenchmini subset.} The bar plot shows the fraction of questions that were answered completely correctly, highlighting the relative performance of different models on both the \chembench corpus and the \chembenchmini subset.
        We see that the model ranking remains fairly consistent across both sets}
    \label{fig:performance_corpus_and_tiny}
    \script{corpus_humanset_performance.py}
\end{figure}

\Cref{fig:performance_corpus_and_tiny} shows the performance of the models on the \chembench corpus and the \chembenchmini subset. The relative performance difference between both is quite similar across most models.
This makes \chembenchmini subset a reliable subset for human baseline comparison and particularly valuable for rapid prototyping and initial model assessment phases.

An interesting observation is the significant impact of chemical preference tasks on \GPTFour's scores. A detailed breakdown of overall accuracy into scores on different skills and difficulty levels is provided in \Cref{tab:performance_table} and \Cref{tab:performance_table_human_subset} for \chembench corpus and the \chembenchmini subset, respectively.

\begin{table}
    \caption{\textbf{Performance of the models on the \chembench corpus.} The table shows the fraction of questions answered correctly by the models for different skills and difficulty levels. Models with \enquote{T-one} in the name were run for a temperature of 1, which allows us to study the temperature effect in the benchmark. Systems with \enquote{ReAct} in the name are tool augmented, i.e., they can call external tools such as web search or a calculator to answer the questions better. However, we limit those systems to ten calls to the \gls{llm}. This constraint often led the systems to not find the correct answer within the specified number of calls. In this case, we consider the answer as incorrect (see \Cref{sec:react-environment}).}
    \resizebox{\textwidth}{!}{
  \begin{tabular}{ccccccccc}
\toprule
\multirow{3}{*}{Model} & \multicolumn{4}{c}{\textbf{Requires}} & \multicolumn{3}{c}{\textbf{Difficulty}} & \multirow{3}{*}{\textbf{Overall Accuracy}}\\\cmidrule(lr){2-5} \cmidrule(lr){6-8}\\
 & Calculation & Knowledge & Reasoning & Intuition & Basic & Intermediate & Advanced &  \\
\midrule
Claude-2 & 0.31 & 0.37 & 0.54 & 0.51 & 0.63 & 0.41 & 0.36 & 0.47 \\
Claude-3 (Opus) & 0.59 & 0.47 & 0.67 & 0.57 & 0.77 & 0.50 & 0.43 & 0.57 \\
Claude-3.5 (Sonnet) & 0.65 & 0.53 & 0.77 & 0.58 & 0.83 & \textbf{0.55} & 0.64 & 0.63 \\
Claude-3.5 (Sonnet) React & 0.71 & 0.50 & 0.77 & \textbf{0.60} & 0.81 & 0.54 & 0.75 & 0.62 \\
Command-R+ & 0.19 & 0.35 & 0.50 & 0.51 & 0.55 & 0.40 & 0.20 & 0.45 \\
Galactica-120b & 0.02 & 0.02 & 0.03 & 0.00 & 0.04 & 0.00 & 0.00 & 0.02 \\
Gemini-Pro & 0.25 & 0.36 & 0.51 & 0.50 & 0.58 & 0.40 & 0.35 & 0.45 \\
Gemma-1.1-7B-it & 0.20 & 0.24 & 0.36 & 0.00 & 0.43 & 0.09 & 0.10 & 0.19 \\
Gemma-1.1-7B-it-T-one & 0.19 & 0.23 & 0.36 & 0.01 & 0.44 & 0.10 & 0.14 & 0.19 \\
Gemma-2-9B-it & 0.40 & 0.36 & 0.52 & 0.55 & 0.61 & 0.43 & 0.38 & 0.48 \\
Gemma-2-9B-it-T-one & 0.37 & 0.35 & 0.51 & 0.56 & 0.61 & 0.43 & 0.43 & 0.48 \\
GPT-3.5 Turbo & 0.25 & 0.35 & 0.53 & 0.53 & 0.60 & 0.42 & 0.36 & 0.47 \\
GPT-4 & 0.48 & 0.46 & 0.65 & 0.16 & 0.74 & 0.28 & 0.57 & 0.41 \\
GPT-4o & 0.72 & 0.51 & 0.73 & 0.59 & 0.81 & 0.53 & 0.62 & 0.61 \\
GPT-4o React & 0.54 & 0.44 & 0.67 & 0.42 & 0.73 & 0.41 & 0.60 & 0.51 \\
Llama-2-70B Chat & 0.10 & 0.14 & 0.15 & 0.49 & 0.20 & 0.30 & 0.00 & 0.27 \\
Llama-3-8B-Instruct & 0.25 & 0.37 & 0.50 & 0.52 & 0.55 & 0.41 & 0.52 & 0.46 \\
Llama-3-8B-Instruct-T-one & 0.23 & 0.36 & 0.51 & 0.52 & 0.55 & 0.42 & 0.57 & 0.46 \\
Llama-3-70B-Instruct & 0.47 & 0.42 & 0.60 & 0.53 & 0.69 & 0.45 & 0.29 & 0.52 \\
Llama-3-70B-Instruct-T-one & 0.46 & 0.42 & 0.59 & 0.53 & 0.70 & 0.45 & 0.29 & 0.52 \\
Llama-3.1-8B-Instruct & 0.34 & 0.36 & 0.52 & 0.53 & 0.61 & 0.41 & 0.38 & 0.47 \\
Llama-3.1-8B-Instruct-T-one & 0.35 & 0.35 & 0.50 & 0.52 & 0.59 & 0.41 & 0.64 & 0.46 \\
Llama-3.1-70B-Instruct & 0.50 & 0.44 & 0.65 & 0.52 & 0.74 & 0.45 & 0.33 & 0.53 \\
Llama-3.1-70B-Instruct-T-one & 0.50 & 0.43 & 0.57 & 0.54 & 0.65 & 0.46 & 0.36 & 0.51 \\
Llama-3.1-405B-Instruct & 0.64 & 0.49 & 0.71 & 0.54 & 0.79 & 0.50 & 0.57 & 0.58 \\
Mistral-Large-2 & 0.65 & 0.47 & 0.70 & 0.55 & 0.78 & 0.48 & 0.57 & 0.57 \\
Mixtral-8x7b-Instruct & 0.21 & 0.31 & 0.42 & 0.54 & 0.51 & 0.39 & 0.25 & 0.42 \\
Mixtral-8x7b-Instruct-T-one & 0.21 & 0.31 & 0.41 & 0.52 & 0.51 & 0.38 & 0.25 & 0.42 \\
o1 & \textbf{0.80} & \textbf{0.57} & \textbf{0.80} & 0.56 & \textbf{0.86} & \textbf{0.55} & \textbf{0.85} & \textbf{0.64} \\
PaperQA2 & 0.62 & 0.49 & 0.66 & 0.56 & 0.75 & 0.49 & 0.57 & 0.57 \\
Phi-3-Medium-4k-Instruct & 0.34 & 0.36 & 0.53 & 0.53 & 0.61 & 0.42 & 0.50 & 0.47 \\
\bottomrule
\end{tabular}
\unskip\label{output/performance_table.tex}\unskip%

    }
    \label{tab:performance_table}
\end{table}

\begin{table}
    \caption{\textbf{Performance of the models on \chembenchmini.} The table shows the fraction of questions answered correctly by the models for different skills and difficulty levels.}
    \resizebox{\textwidth}{!}{
  \begin{tabular}{cccccccc}
\toprule
\multirow{3}{*}{Model} & \multicolumn{3}{c}{\textbf{Requires}} & \multicolumn{3}{c}{\textbf{Difficulty}} & \multirow{3}{*}{\textbf{Overall Accuracy}}\\\cmidrule(lr){2-4} \cmidrule(lr){5-7}\\
 & Calculation & Knowledge & Reasoning & Basic & Intermediate & Advanced &  \\
\midrule
Claude-2 & 0.29 & 0.52 & 0.51 & 0.57 & 0.45 & 0.36 & 0.49 \\
Claude-3 (Opus) & 0.62 & 0.64 & 0.67 & 0.78 & 0.61 & 0.43 & 0.67 \\
Claude-3.5 (Sonnet) & 0.62 & 0.75 & 0.75 & 0.79 & 0.70 & 0.64 & 0.73 \\
Claude-3.5 (Sonnet) React & 0.73 & 0.75 & 0.75 & \textbf{0.83} & 0.71 & 0.75 & 0.76 \\
Command-R+ & 0.25 & 0.49 & 0.41 & 0.46 & 0.42 & 0.20 & 0.42 \\
Galactica-120b & 0.01 & 0.02 & 0.01 & 0.04 & 0.01 & 0.00 & 0.02 \\
Gemini-Pro & 0.27 & 0.49 & 0.46 & 0.50 & 0.43 & 0.35 & 0.45 \\
Gemma-1.1-7B-it & 0.23 & 0.40 & 0.36 & 0.42 & 0.33 & 0.10 & 0.35 \\
Gemma-1.1-7B-it-T-one & 0.21 & 0.39 & 0.35 & 0.41 & 0.31 & 0.14 & 0.34 \\
Gemma-2-9B-it & 0.44 & 0.46 & 0.45 & 0.56 & 0.44 & 0.40 & 0.49 \\
Gemma-2-9B-it-T-one & 0.40 & 0.44 & 0.42 & 0.55 & 0.41 & 0.43 & 0.46 \\
GPT-3.5 Turbo & 0.31 & 0.46 & 0.45 & 0.51 & 0.41 & 0.36 & 0.44 \\
GPT-4 & 0.47 & 0.63 & 0.64 & 0.67 & 0.62 & 0.57 & 0.64 \\
GPT-4o & 0.68 & 0.72 & 0.74 & 0.81 & 0.67 & 0.65 & 0.72 \\
GPT-4o React & 0.60 & 0.59 & 0.63 & 0.69 & 0.57 & 0.60 & 0.62 \\
Human & 0.28 & 0.23 & 0.27 & 0.32 & 0.24 & 0.27 & 0.27 \\
Llama-2-70B Chat & 0.12 & 0.18 & 0.13 & 0.19 & 0.13 & 0.00 & 0.14 \\
Llama-3-8B-Instruct & 0.26 & 0.54 & 0.44 & 0.43 & 0.45 & 0.50 & 0.44 \\
Llama-3-8B-Instruct-T-one & 0.28 & 0.54 & 0.45 & 0.43 & 0.46 & 0.57 & 0.45 \\
Llama-3-70B-Instruct & 0.49 & 0.63 & 0.57 & 0.67 & 0.59 & 0.30 & 0.60 \\
Llama-3-70B-Instruct-T-one & 0.49 & 0.61 & 0.57 & 0.67 & 0.57 & 0.29 & 0.59 \\
Llama-3.1-8B-Instruct & 0.33 & 0.47 & 0.44 & 0.55 & 0.39 & 0.40 & 0.46 \\
Llama-3.1-8B-Instruct-T-one & 0.38 & 0.47 & 0.40 & 0.51 & 0.38 & 0.64 & 0.44 \\
Llama-3.1-70B-Instruct & 0.52 & 0.62 & 0.63 & 0.73 & 0.62 & 0.35 & 0.64 \\
Llama-3.1-70B-Instruct-T-one & 0.50 & 0.55 & 0.58 & 0.66 & 0.54 & 0.36 & 0.57 \\
Llama-3.1-405B-Instruct & 0.66 & 0.69 & 0.70 & 0.72 & 0.69 & 0.57 & 0.69 \\
Mistral-Large-2 & 0.66 & 0.65 & 0.68 & 0.72 & 0.66 & 0.60 & 0.68 \\
Mixtral-8x7b-Instruct & 0.23 & 0.40 & 0.37 & 0.45 & 0.32 & 0.25 & 0.36 \\
Mixtral-8x7b-Instruct-T-one & 0.21 & 0.39 & 0.35 & 0.46 & 0.29 & 0.25 & 0.36 \\
o1 & \textbf{0.77} & \textbf{0.79} & \textbf{0.81} & 0.82 & \textbf{0.77} & \textbf{0.85} & \textbf{0.80} \\
PaperQA2 & 0.63 & 0.69 & 0.65 & 0.70 & 0.67 & 0.55 & 0.67 \\
Phi-3-Medium-4k-Instruct & 0.34 & 0.49 & 0.52 & 0.57 & 0.47 & 0.50 & 0.51 \\
\bottomrule
\end{tabular}
\unskip\label{output/performance_table_human_subset.tex}\unskip%

    }
    \label{tab:performance_table_human_subset}
\end{table}

\begin{table}
    \caption{\textbf{Performance of the models on the \chembench corpus.} The table shows the fraction of questions answered correctly by the models for different topics. Models with \enquote{T-one} in the name were run for a temperature of 1, which allows us to study the temperature effect in the benchmark. Systems with \enquote{ReAct} in the name are tool augmented, i.e., they can call external tools such as web search or a calculator to answer the questions better. However, we limit those systems to ten calls to the \gls{llm}. This constraint often led the systems to not find the correct answer within the specified number of calls. In this case, we consider the answer as incorrect (see \Cref{sec:react-environment}).}
    \resizebox{\textwidth}{!}{
  \begin{tabular}{lllllllllll}
\toprule
Model & Analytical & Chemical Preference & General & Inorganic & Materials Science & Organic & Physical & Technical & Toxicity/Safety & Overall Accuracy \\
\midrule
Claude-2 & 0.39 & 0.51 & 0.51 & 0.56 & 0.30 & 0.60 & 0.54 & 0.47 & 0.33 & 0.47 \\
Claude-3 (Opus) & 0.48 & 0.57 & 0.77 & 0.75 & 0.50 & 0.69 & 0.69 & 0.70 & 0.41 & 0.57 \\
Claude-3.5 (Sonnet) & 0.57 & 0.58 & 0.83 & 0.82 & 0.50 & 0.82 & 0.81 & \textbf{0.85} & 0.44 & 0.63 \\
Claude-3.5 (Sonnet) React & 0.58 & \textbf{0.60} & 0.87 & 0.79 & 0.50 & \textbf{0.83} & 0.80 & 0.80 & 0.41 & 0.62 \\
Command-R+ & 0.36 & 0.51 & 0.49 & 0.50 & 0.30 & 0.56 & 0.37 & 0.50 & 0.31 & 0.45 \\
Galactica-120b & 0.00 & 0.00 & 0.05 & 0.05 & 0.00 & 0.00 & 0.06 & 0.00 & 0.02 & 0.02 \\
Gemini-Pro & 0.41 & 0.50 & 0.49 & 0.44 & 0.50 & 0.58 & 0.46 & 0.47 & 0.31 & 0.45 \\
Gemma-1.1-7B-it & 0.21 & 0.00 & 0.33 & 0.43 & 0.40 & 0.39 & 0.34 & 0.38 & 0.23 & 0.19 \\
Gemma-1.1-7B-it-T-one & 0.21 & 0.01 & 0.35 & 0.41 & 0.40 & 0.38 & 0.35 & 0.38 & 0.22 & 0.19 \\
Gemma-2-9B-it & 0.32 & 0.55 & 0.55 & 0.53 & 0.50 & 0.57 & 0.54 & 0.53 & 0.34 & 0.48 \\
Gemma-2-9B-it-T-one & 0.30 & 0.56 & 0.56 & 0.51 & 0.40 & 0.57 & 0.53 & 0.47 & 0.34 & 0.48 \\
GPT-3.5 Turbo & 0.40 & 0.53 & 0.49 & 0.53 & 0.50 & 0.60 & 0.44 & 0.40 & 0.31 & 0.47 \\
GPT-4 & 0.44 & 0.16 & 0.70 & 0.68 & 0.50 & 0.68 & 0.69 & 0.70 & 0.41 & 0.41 \\
GPT-4o & 0.56 & 0.59 & 0.81 & 0.80 & 0.60 & 0.75 & 0.76 & 0.75 & 0.44 & 0.61 \\
GPT-4o React & 0.47 & 0.42 & 0.76 & 0.73 & 0.40 & 0.72 & 0.62 & 0.72 & 0.37 & 0.51 \\
Llama-2-70B Chat & 0.07 & 0.49 & 0.13 & 0.23 & 0.20 & 0.15 & 0.18 & 0.12 & 0.14 & 0.27 \\
Llama-3-8B-Instruct & 0.43 & 0.52 & 0.45 & 0.48 & 0.40 & 0.57 & 0.40 & 0.60 & 0.32 & 0.46 \\
Llama-3-8B-Instruct-T-one & 0.42 & 0.52 & 0.44 & 0.52 & 0.30 & 0.55 & 0.38 & 0.62 & 0.32 & 0.46 \\
Llama-3-70B-Instruct & 0.43 & 0.53 & 0.61 & 0.64 & 0.60 & 0.65 & 0.64 & 0.62 & 0.37 & 0.52 \\
Llama-3-70B-Instruct-T-one & 0.39 & 0.53 & 0.61 & 0.66 & 0.60 & 0.64 & 0.65 & 0.60 & 0.37 & 0.52 \\
Llama-3.1-8B-Instruct & 0.41 & 0.53 & 0.51 & 0.50 & 0.40 & 0.58 & 0.55 & 0.45 & 0.33 & 0.47 \\
Llama-3.1-8B-Instruct-T-one & 0.38 & 0.52 & 0.53 & 0.48 & 0.40 & 0.58 & 0.48 & 0.40 & 0.32 & 0.46 \\
Llama-3.1-70B-Instruct & 0.42 & 0.52 & 0.69 & 0.74 & 0.50 & 0.68 & 0.68 & 0.65 & 0.38 & 0.53 \\
Llama-3.1-70B-Instruct-T-one & 0.36 & 0.54 & 0.67 & 0.67 & 0.50 & 0.58 & 0.55 & 0.55 & 0.39 & 0.51 \\
Llama-3.1-405B-Instruct & 0.52 & 0.54 & 0.79 & 0.78 & 0.70 & 0.74 & 0.73 & 0.70 & 0.42 & 0.58 \\
Mistral-Large-2 & 0.50 & 0.55 & 0.79 & 0.77 & 0.40 & 0.73 & 0.73 & 0.68 & 0.40 & 0.57 \\
Mixtral-8x7b-Instruct & 0.28 & 0.54 & 0.43 & 0.51 & 0.40 & 0.50 & 0.37 & 0.33 & 0.27 & 0.42 \\
Mixtral-8x7b-Instruct-T-one & 0.28 & 0.52 & 0.45 & 0.50 & 0.40 & 0.50 & 0.39 & 0.33 & 0.27 & 0.42 \\
o1 & \textbf{0.62} & 0.56 & \textbf{0.93} & \textbf{0.90} & \textbf{0.80} & 0.82 & \textbf{0.89} & \textbf{0.85} & \textbf{0.48} & \textbf{0.64} \\
PaperQA2 & 0.48 & 0.56 & 0.73 & 0.71 & 0.60 & 0.67 & 0.73 & 0.70 & 0.42 & 0.57 \\
Phi-3-Medium-4k-Instruct & 0.35 & 0.53 & 0.48 & 0.58 & 0.40 & 0.56 & 0.49 & 0.55 & 0.33 & 0.47 \\
\bottomrule
\end{tabular}
\unskip\label{output/performance_topic_table.tex}\unskip%

    }
    \label{tab:performance_table_topic}
\end{table}

\begin{table}
    \caption{\textbf{Performance of the models on \chembenchmini.} The table shows the fraction of questions answered correctly by the models for different topics.}
    \resizebox{\textwidth}{!}{
  \begin{tabular}{cccccccccc}
\toprule
Model & Analytical & General & Inorganic & Materials Science & Organic & Physical & Technical & Toxicity/Safety & Overall Accuracy \\
\midrule
Claude-2 & 0.77 & 0.46 & 0.47 & 0.00 & 0.62 & 0.42 & 0.52 & 0.60 & 0.49 \\
Claude-3 (Opus) & 0.77 & 0.73 & 0.76 & 0.50 & 0.62 & 0.62 & 0.74 & 0.65 & 0.67 \\
Claude-3.5 (Sonnet) & 0.86 & 0.69 & 0.68 & 0.50 & 0.79 & \textbf{0.79} & 0.87 & 0.55 & 0.73 \\
Claude-3.5 (Sonnet) React & 0.77 & 0.81 & 0.71 & 0.50 & \textbf{0.88} & \textbf{0.79} & 0.83 & 0.55 & 0.76 \\
Command-R+ & 0.45 & 0.35 & 0.44 & 0.00 & 0.33 & 0.25 & 0.57 & 0.60 & 0.42 \\
Galactica-120b & 0.00 & 0.00 & 0.03 & 0.00 & 0.00 & 0.12 & 0.00 & 0.05 & 0.02 \\
Gemini-Pro & 0.55 & 0.38 & 0.41 & 0.00 & 0.58 & 0.46 & 0.43 & 0.45 & 0.45 \\
Gemma-1.1-7B-it & 0.36 & 0.31 & 0.44 & 0.50 & 0.46 & 0.29 & 0.39 & 0.40 & 0.35 \\
Gemma-1.1-7B-it-T-one & 0.32 & 0.31 & 0.41 & 0.50 & 0.42 & 0.33 & 0.39 & 0.40 & 0.34 \\
Gemma-2-9B-it & 0.45 & 0.54 & 0.47 & 0.50 & 0.54 & 0.58 & 0.61 & 0.50 & 0.49 \\
Gemma-2-9B-it-T-one & 0.45 & 0.50 & 0.44 & 0.50 & 0.58 & 0.54 & 0.52 & 0.45 & 0.46 \\
GPT-3.5 Turbo & 0.55 & 0.35 & 0.44 & 0.50 & 0.58 & 0.50 & 0.43 & 0.55 & 0.44 \\
GPT-4 & 0.77 & 0.69 & 0.59 & \textbf{1.00} & 0.79 & 0.67 & 0.83 & 0.60 & 0.64 \\
GPT-4o & 0.73 & 0.69 & 0.71 & 0.50 & 0.75 & 0.75 & 0.78 & 0.60 & 0.72 \\
GPT-4o React & 0.55 & 0.58 & 0.68 & 0.50 & 0.75 & 0.58 & 0.74 & 0.45 & 0.62 \\
Human & 0.31 & 0.41 & 0.30 & 0.24 & 0.36 & 0.26 & 0.20 & 0.22 & 0.27 \\
Llama-2-70B Chat & 0.00 & 0.19 & 0.15 & 0.50 & 0.21 & 0.17 & 0.13 & 0.25 & 0.14 \\
Llama-3-8B-Instruct & 0.68 & 0.27 & 0.44 & 0.00 & 0.50 & 0.29 & 0.61 & 0.50 & 0.44 \\
Llama-3-8B-Instruct-T-one & 0.64 & 0.27 & 0.53 & 0.00 & 0.46 & 0.25 & 0.65 & 0.50 & 0.45 \\
Llama-3-70B-Instruct & 0.64 & 0.62 & 0.62 & 0.50 & 0.75 & 0.58 & 0.74 & 0.50 & 0.60 \\
Llama-3-70B-Instruct-T-one & 0.59 & 0.62 & 0.62 & 0.50 & 0.71 & 0.58 & 0.70 & 0.60 & 0.59 \\
Llama-3.1-8B-Instruct & 0.59 & 0.35 & 0.53 & 0.00 & 0.50 & 0.50 & 0.52 & 0.60 & 0.46 \\
Llama-3.1-8B-Instruct-T-one & 0.59 & 0.38 & 0.47 & 0.50 & 0.50 & 0.42 & 0.39 & 0.55 & 0.44 \\
Llama-3.1-70B-Instruct & 0.59 & 0.62 & 0.74 & 0.50 & 0.75 & 0.71 & 0.70 & 0.60 & 0.64 \\
Llama-3.1-70B-Instruct-T-one & 0.45 & 0.58 & 0.59 & 0.50 & 0.67 & 0.62 & 0.65 & 0.45 & 0.57 \\
Llama-3.1-405B-Instruct & 0.68 & 0.73 & 0.74 & \textbf{1.00} & 0.71 & 0.75 & 0.70 & 0.60 & 0.69 \\
Mistral-Large-2 & 0.82 & 0.65 & 0.68 & \textbf{1.00} & 0.75 & 0.71 & 0.70 & 0.50 & 0.68 \\
Mixtral-8x7b-Instruct & 0.41 & 0.35 & 0.47 & 0.50 & 0.33 & 0.33 & 0.30 & 0.40 & 0.36 \\
Mixtral-8x7b-Instruct-T-one & 0.45 & 0.42 & 0.41 & 0.50 & 0.38 & 0.38 & 0.30 & 0.35 & 0.36 \\
o1 & \textbf{0.91} & \textbf{0.92} & \textbf{0.79} & \textbf{1.00} & 0.79 & \textbf{0.79} & \textbf{0.91} & \textbf{0.70} & \textbf{0.80} \\
PaperQA2 & 0.68 & 0.73 & 0.62 & \textbf{1.00} & 0.58 & 0.71 & 0.78 & 0.65 & 0.67 \\
Phi-3-Medium-4k-Instruct & 0.45 & 0.42 & 0.56 & 0.50 & 0.62 & 0.54 & 0.57 & 0.50 & 0.51 \\
\bottomrule
\end{tabular}
\unskip\label{output/performance_topic_table_human_subset.tex}\unskip%

    }
    \label{tab:performance_table_human_subset_topic}
\end{table}

\clearpage

\subsection{Performance as a function of molecular features} \label{sec:molecular_features}
\begin{figure}[!h]
    \centering
    \includegraphics[width=\textwidth]{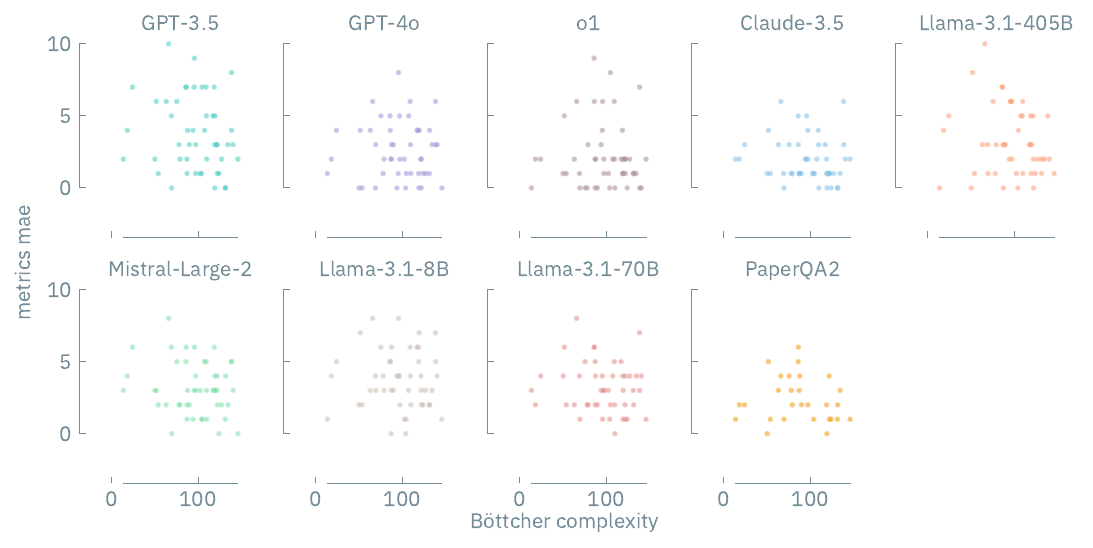}
    \caption{\textbf{Dependence of the mean absolute error in predicting the number of NMR signals on the Böttcher complexity of the molecules.} The complexity measure proposed by \textcite{B_ttcher_2016} is an information-theoretic additive measure of compound complexity that follows chemical intuitions.
    The plot shows that for the \glspl{llm}, the predictive performance (measured as the mean absolute error in the prediction of the number of \gls{nmr} signals) is not correlated with the complexity of the molecules (that is, molecules tend to not to be able to predict the number of \gls{nmr} signals regardless of molecular complexity). For inference based on reasoning, one would expect that the complexity of the molecule is a good predictor of the difficulty of the question.}
    \script{correlate_with_molecule_features.py}
    \label{fig:correlation_plot_is_number_nmr_peaks_complexity}
\end{figure}

\begin{figure}[!h]
    \centering
    \includegraphics[width=\textwidth]{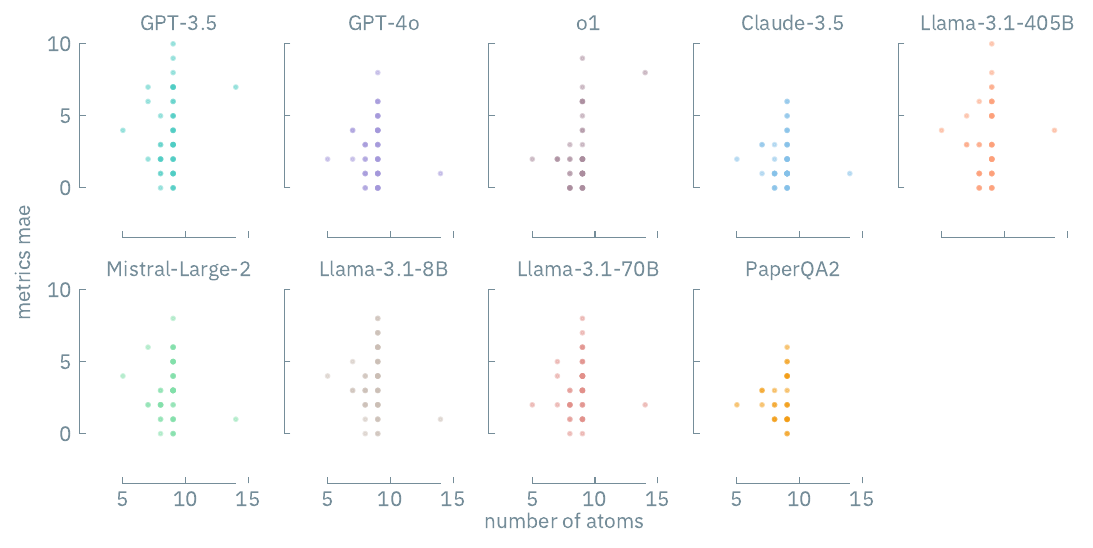}
    \caption{\textbf{Dependence of the mean absolute error in predicting the number of NMR signals on the number of atoms.} }
    \script{correlate_with_molecule_features.py}
    \label{fig:correlation_plot_is_number_nmr_peaks_num_atoms}
\end{figure}

\begin{figure}[!h]
    \centering
    \includegraphics[width=\textwidth]{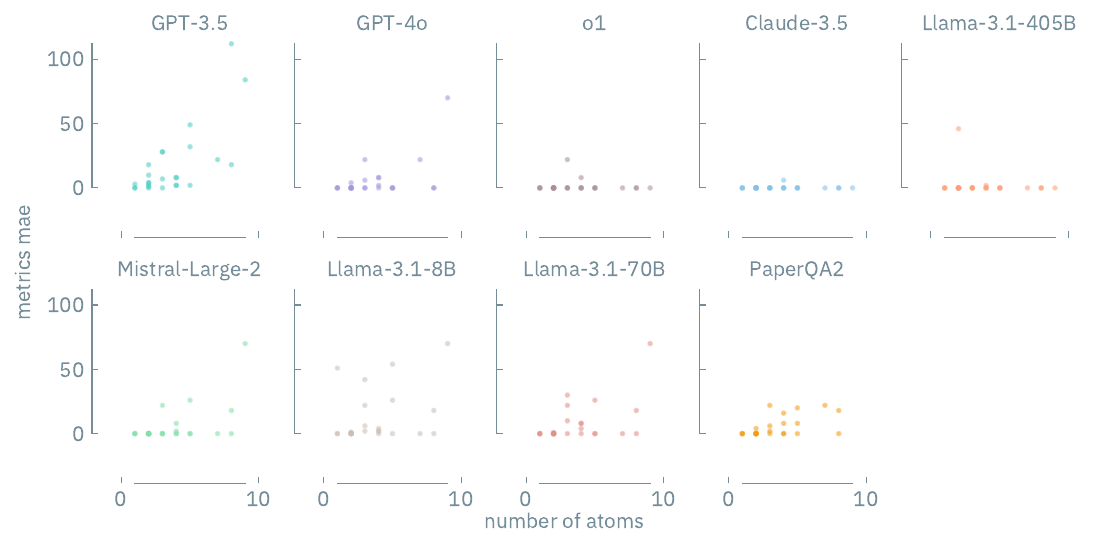}
    \caption{\textbf{Dependence of the mean absolute error in predicting total electron counts on the number of atoms.} The plot shows that for the \glspl{llm}, the predictive performance (measured as the mean absolute error in the prediction of the total electron counts) is sometimes correlated with the number of atoms in the molecule.}
    \script{correlate_with_molecule_features.py}
    \label{fig:correlation_plot_is_electron_counts_num_atoms}
\end{figure}

To better understand if the performance of the models is correlated with specific features of the molecules, we analyzed the performance of the models as a function of the number of atoms and the complexity of the molecules.
\Cref{fig:correlation_plot_is_number_nmr_peaks_complexity} shows that the performance of the models is not correlated with the complexity of the molecules but rather with the number of atoms (\Cref{fig:correlation_plot_is_number_nmr_peaks_num_atoms}, similar trivial correlation for \Cref{fig:correlation_plot_is_electron_counts_num_atoms}).

\clearpage
\subsection{Influence of model scale}
\begin{figure}[!h]
    \centering
    \includegraphics[width=\textwidth]{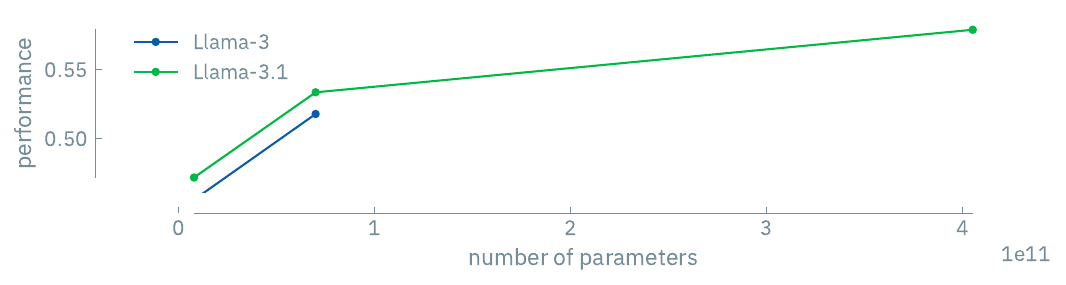}
    \caption{\textbf{Performance of models as a function of model size.} The plot shows the performance of the models as a function of the parameter count. The performance is measured as the fraction of questions answered correctly by the models. We see that the performance of the models correlates with scale for the models of the LLama-3 and Llama-3.1 herd of models.}
    \script{performance_vs_model_size.py}
    \label{fig:model_size_plot}
\end{figure}

\Cref{fig:model_size_plot} shows the performance of the models as a function of the number of parameters in the model.
We see that the performance of the models correlates with their size for the models of the LLama-3 and Llama-3.1 herd of models.

\clearpage
\subsection{Refusal detection}
\Glspl{llm} typically undergo refusal training to prevent harmful or undesirable outputs. As a result, models may decline to answer questions perceived as potentially adversarial prompts.
To automatically detect refusals, we use a modified regular expression reported by LLM Guard\autocite{llmguard} to detect commonly used refusal phrases.

\Cref{tab:refusal_counts_and_parsing} lists how many refusals were detected for the responses of different models on \chembench. Overall, we find that refusals do not majorly affect the performance measured by \chembench.

\subsection{LLM Parsing} \label{sec:llm-parsing}

Our parsing workflow uses pipelines based on regular expressions to extract the answers. In some cases, however, the answers are not directly extractable from the responses, for instance, when the model does not follow the formatting instructions. In these cases, we use a fallback mechanism to extract the answers. The fallback mechanism uses an \gls{llm} to extract the answers from the responses. The \gls{llm} is provided with the response and the question and is prompted to only extract but not generate the answer. We used \LlamaThreeSeventyBInstruct, accessed via the Groq API.
\Cref{tab:refusal_counts_and_parsing} tabulates the number of times the fallback mechanism was used for each model.

\begin{table}
    \centering
    \caption{\textbf{Refusal counts and parsing.} The table shows the number of refusals detected and the number of times the \gls{llm} fallback parsing mechanism was used for each model.}
  \begin{tabular}{lcccc}
\toprule
\multirow{3}{*}{Model} & \multicolumn{2}{c}{\textbf{Refusal}} & \multicolumn{2}{c}{\textbf{LLM Extraction}}\\
\cmidrule(lr){2-3} \cmidrule(lr){4-5}\\
& Nº of Questions & Fraction & Nº of Questions & Fraction\\
\midrule
Claude 2 & 100 & 0.035868 & 0 & 0.000000 \\
Claude 3 & 9 & 0.003228 & 0 & 0.000000 \\
Claude 3.5 Sonnet & 31 & 0.011119 & 3 & 0.001076 \\
Claude-3.5 ReAct & 0 & 0.000000 & 129 & 0.046270 \\
Command R+ & 0 & 0.000000 & 5 & 0.001793 \\
Gemini Pro & 0 & 0.000000 & 4 & 0.001435 \\
Gemma 1.1 7B & 0 & 0.000000 & 18 & 0.006456 \\
Gemma 1.1 7B Temp=1 & 1 & 0.000359 & 22 & 0.007891 \\
Gemma 2 9B & 5 & 0.001793 & 25 & 0.008967 \\
Gemma 2 9B Temp=1 & 5 & 0.001793 & 37 & 0.013271 \\
GPT-3.5 Turbo & 0 & 0.000000 & 8 & 0.002869 \\
GPT-4 & 0 & 0.000000 & 3 & 0.001076 \\
GPT-4o & 0 & 0.000000 & 7 & 0.002511 \\
GPT-4o ReAct & 0 & 0.000000 & 667 & 0.239240 \\
Llama 2 70B Chat & 1 & 0.000359 & 3 & 0.001076 \\
Llama 3 8B & 0 & 0.000000 & 9 & 0.003228 \\
Llama 3 8B Temp=1 & 6 & 0.002152 & 11 & 0.003945 \\
Llama 3 70B & 0 & 0.000000 & 16 & 0.005739 \\
Llama 3 70B Temp=1 & 1 & 0.000359 & 17 & 0.006098 \\
Llama 3.1 8B & 0 & 0.000000 & 35 & 0.012554 \\
Llama 3.1 8B Temp=1 & 3 & 0.001076 & 30 & 0.010760 \\
Llama 3.1 70B & 0 & 0.000000 & 53 & 0.019010 \\
Llama 3.1 70B Temp=1 & 1 & 0.000359 & 245 & 0.087877 \\
Llama 3.1 405B & 0 & 0.000000 & 22 & 0.007891 \\
Mistral Large 2 123B & 0 & 0.000000 & 2 & 0.000717 \\
Mixtral 8x7B & 4 & 0.001435 & 54 & 0.019369 \\
Mixtral 8x7B Temp=1 & 7 & 0.002511 & 58 & 0.020803 \\
o1 & 0 & 0.000000 & 2 & 0.000717 \\
Paper QA & 35 & 0.012554 & 52 & 0.018651 \\
Phi 3 Medium 4K & 0 & 0.000000 & 7 & 0.002511 \\
\bottomrule
\end{tabular}
\unskip\label{output/model_refusal_table.tex}\unskip%

    \label{tab:refusal_counts_and_parsing}
\end{table}

\clearpage
\subsection{Implementation}
\begin{figure}
    \centering
    \includegraphics[height=.7\textheight]{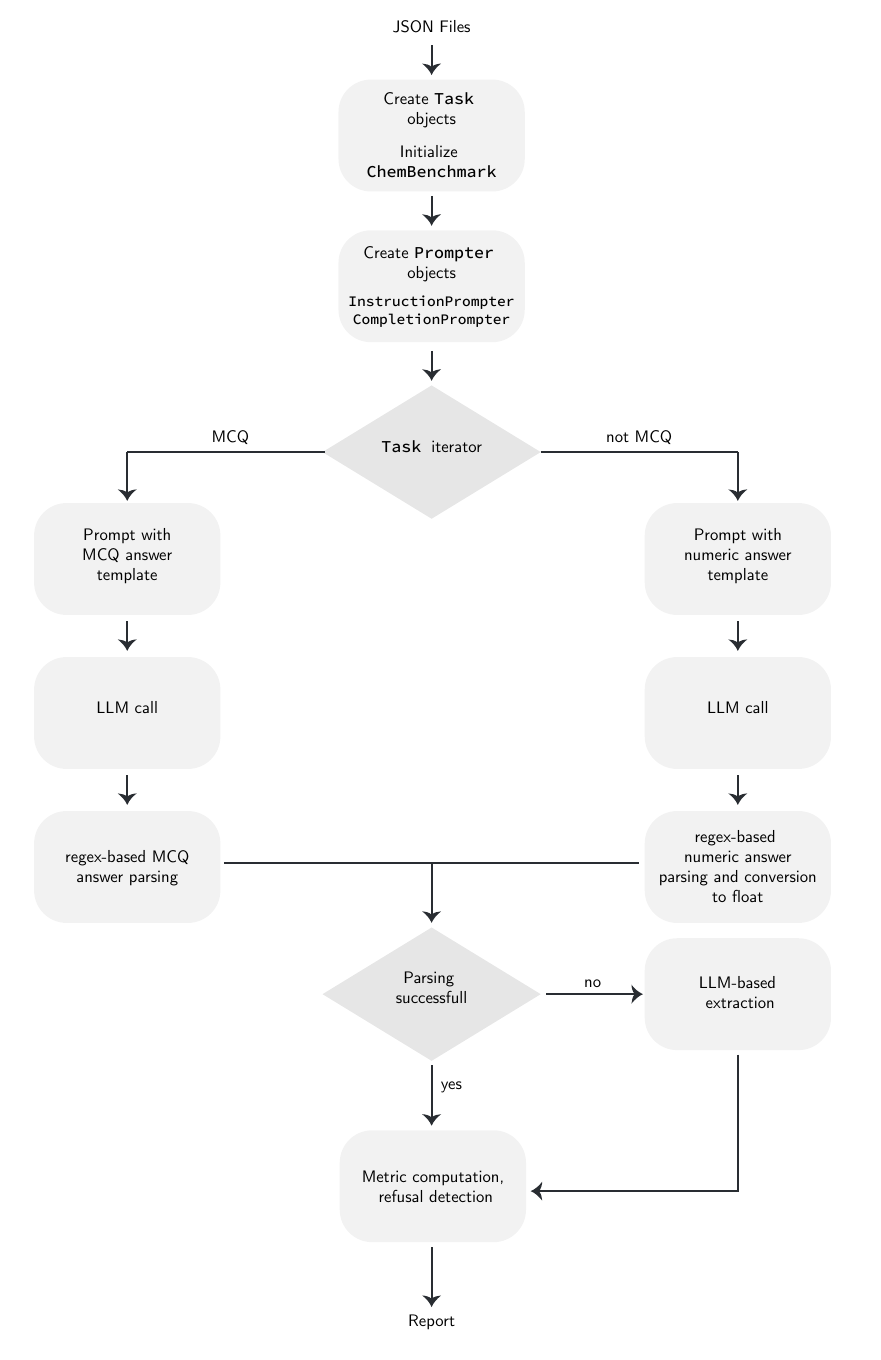}
    \caption{\textbf{Overview of the benchmarking pipeline implemented in \chembench.} The process begins with JSON files containing task data, which are used to create \texttt{Task} objects and initialize the \texttt{ChemBenchmark}. \texttt{Prompter} objects are then created to handle different types of prompts for instruction-tuned and completion models.
    The \texttt{Task} Iterator differentiates between \gls{mcq} and other question types. For each task type, appropriate prompts are generated and passed to the \gls{llm}. The responses are then processed using regex-based parsing methods specific to \gls{mcq} or numeric answers (after obtaining the relevant part of the response from the instruction-tuned models).
    The regex-based parsing is elaborate and can handle special cases such as scientific notation or Roman numerals.
    If the initial parsing is unsuccessful, the system employs an \gls{llm}-based extraction method as a fallback. The parsed or extracted answers then undergo metric computation and refusal detection.}
    \label{fig:process}
\end{figure}

An overview of the benchmarking pipeline implemented in \chembench is shown in \Cref{fig:process}. More detailed information can be found in the online documentation of the \chembench package at \url{https://lamalab-org.github.io/chem-bench/}.

\clearpage
\subsection{Human baseline} \label{sec:human_baseline}
\paragraph{App} To facilitate the collection of responses, we developed a responsive web application in Typescript using the Next.js\autocite{nextjs} app router framework.
This application handles serving the user interface and exposes various \gls{rest} \glspl{api} for relevant operations.
We utilize a Postgresql.
The web application is styled with Tailwind CSS\autocite{tailwindcss} using the shadcn/ui component library and uses NextAuth\autocite{nextauth} for easy and secure user authentication.
The application is hosted on the Vercel web hosting platform.

\begin{figure}
    \subfloat[A physcial chemistry question.]{
        \includegraphics[width=\textwidth]{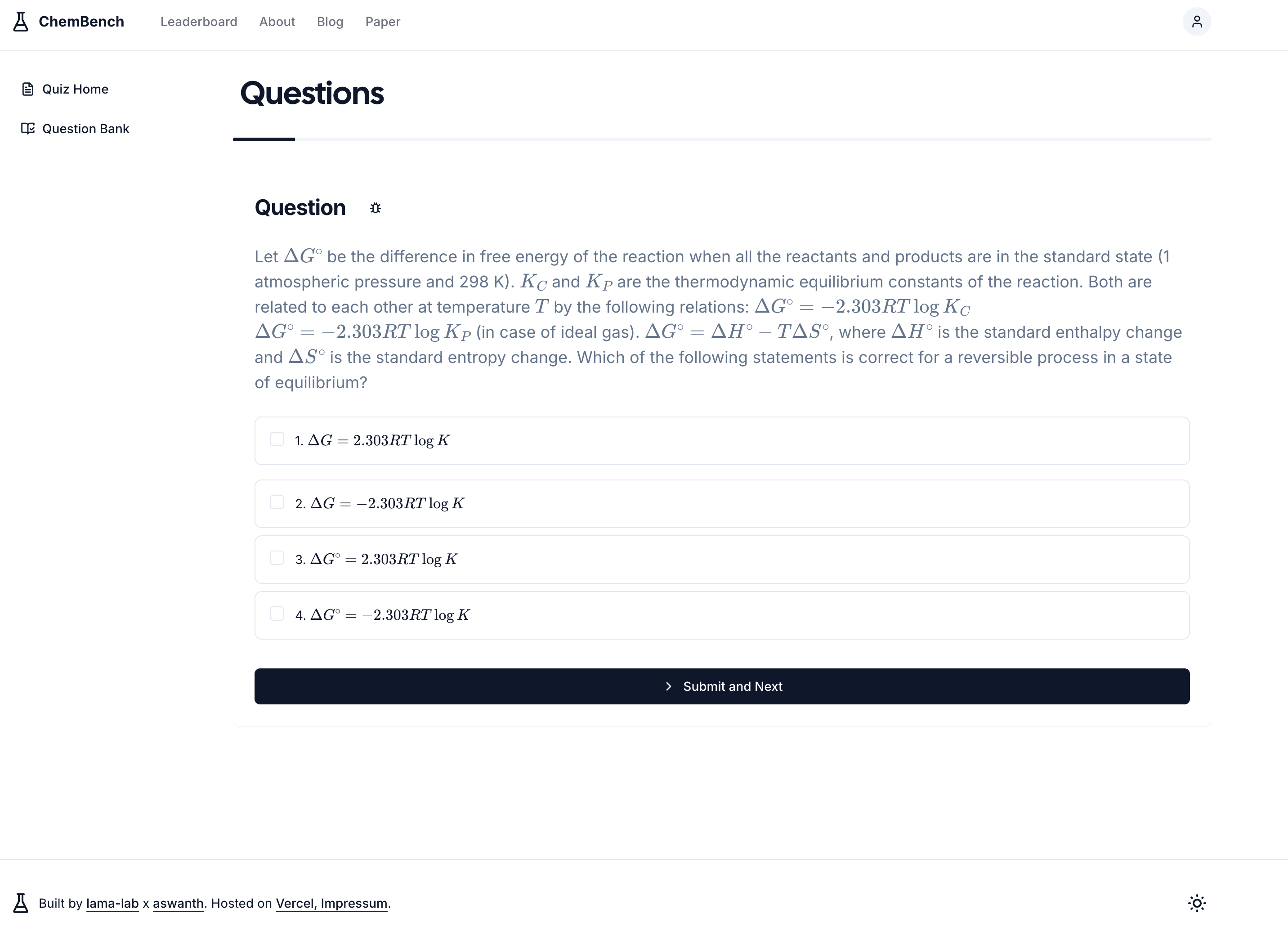}
    }

    \subfloat[An organic chemistry question.]{
        \includegraphics[width=\textwidth]{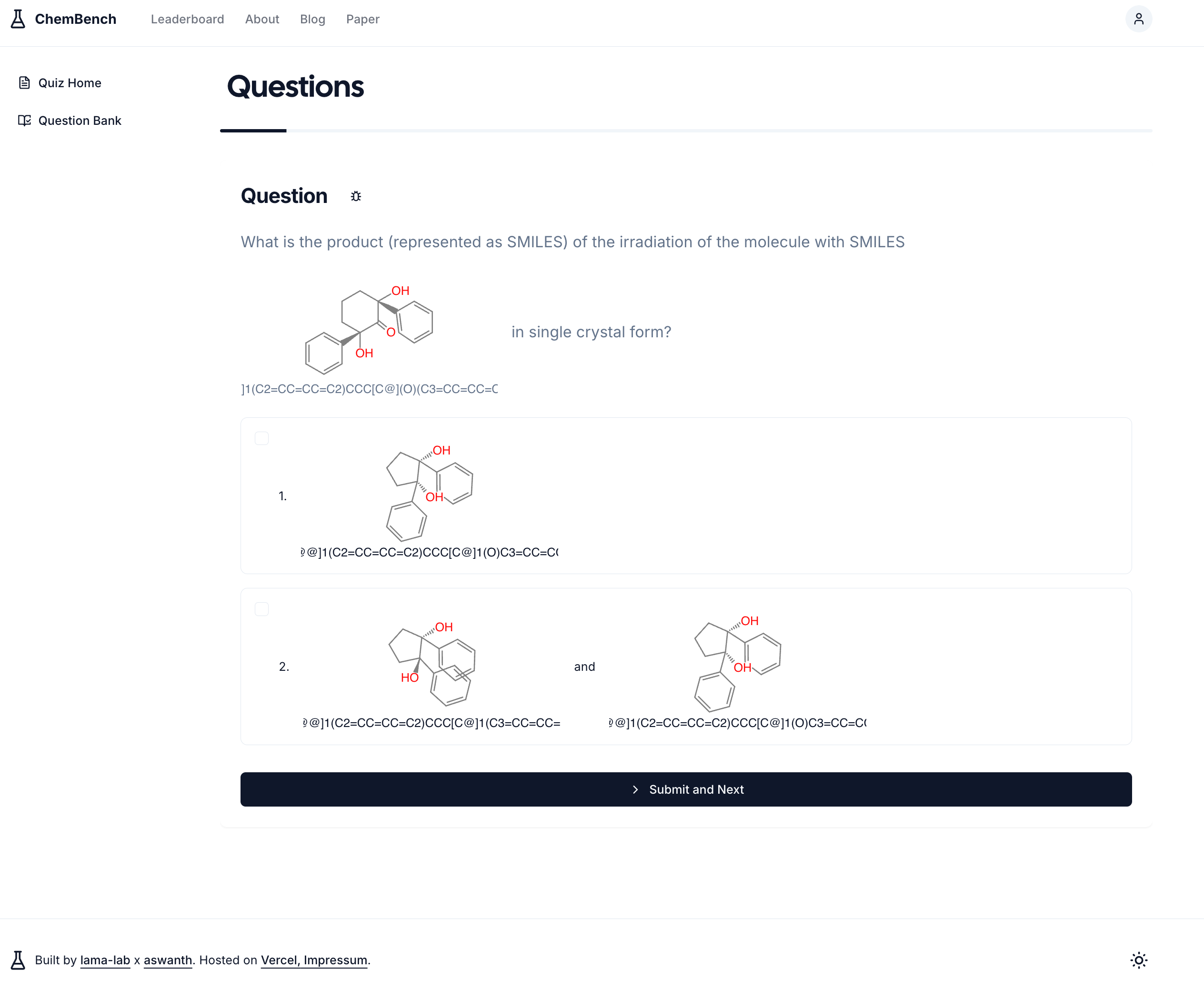}
    }
    \caption{\textbf{Examples of how questions were shown to the human participants.}}
    \label{fig:screenshots}
\end{figure}

In the applications, human participants were presented with molecules as rendered drawings and SMILES strings. \LaTeX\xspace equations and chemical equations were rendered using MathJax (\Cref{fig:screenshots}).

\begin{figure}[htb]
    \centering
    \includegraphics{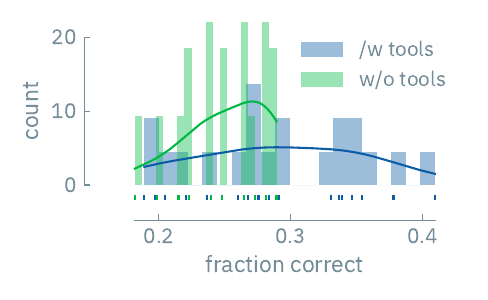}
    \script{plot_human_score_distribution.py}
    \caption{\textbf{Distribution of human scores.}The histogram and kernel density estimates show the fraction of questions answered correctly by the human volunteers.
    Since the best possible score for each question is one and the worst possible score is zero, the values on this plot are between zero and one. A score of one would mean that a volunteer answered all questions correctly. A score of zero would mean that no question was answered correctly.
    We find that the scores for the questions that the human volunteers answered with tools are generally lower than the scores for the questions that the human volunteers answered without tools.}
    \label{fig:human_score_distribution}
\end{figure}

\paragraph{Statistics}
\Cref{fig:human_score_distribution} shows the distribution of scores our human scorers achieved.

\begin{figure}[htb]
    \centering
    \includegraphics{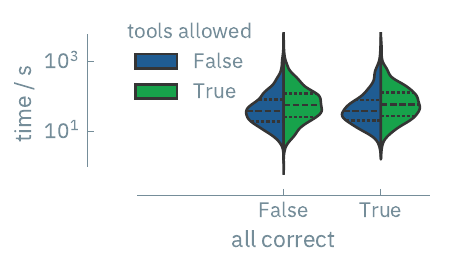}
    \script{analyze_human_data.py}
    \caption{\textbf{Time taken by human scorers to answer questions vs.\ correctness of their answers.} From the plot, it is clear that there is no clear dependence of the correctness of the answers on the time taken by the human scorers to answer the questions. However, we see that human scorers typically took longer to correctly answer questions with tool use.}
    \label{fig:human_timing}
\end{figure}

We also recorded the time humans took to answer the questions (\Cref{fig:human_timing}). This time is the time from the question being displayed to the human to the human submitting the answer.

\begin{figure}[htb]
    \centering
    \includegraphics{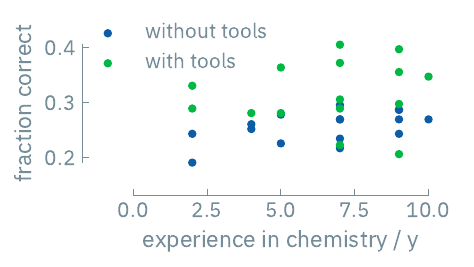}
    \script{analyze_human_data.py}
    \caption{\textbf{Experience of human  scorers vs.\ correctness of their answers.} The experience (in the number of years since the first university-level chemistry course) of the human scorers wasp correlated with the correctness of their answers.}
    \label{fig:experience_vs_correctness}
\end{figure}

Additionally, we prompted users to provide additional information about their experience in chemistry.
While we recorded fine-grained information, e.g., their specialization, we focused on the number of years since the first university-level chemistry course.
\Cref{fig:experience_vs_correctness} shows that the experience of the human scorers was weakly correlated with the correctness of their answers (\Cref{fig:experience_vs_correctness}, Spearman's \(\rho \approx %
  0.19 \unskip\label{output/spearman_experience_score.txt}\unskip%
\), and \(p \approx %
  0.31 \unskip\label{output/spearman_experience_score_p.txt}\unskip%
\)).

\begin{figure}
    \centering
    \includegraphics[width=\textwidth]{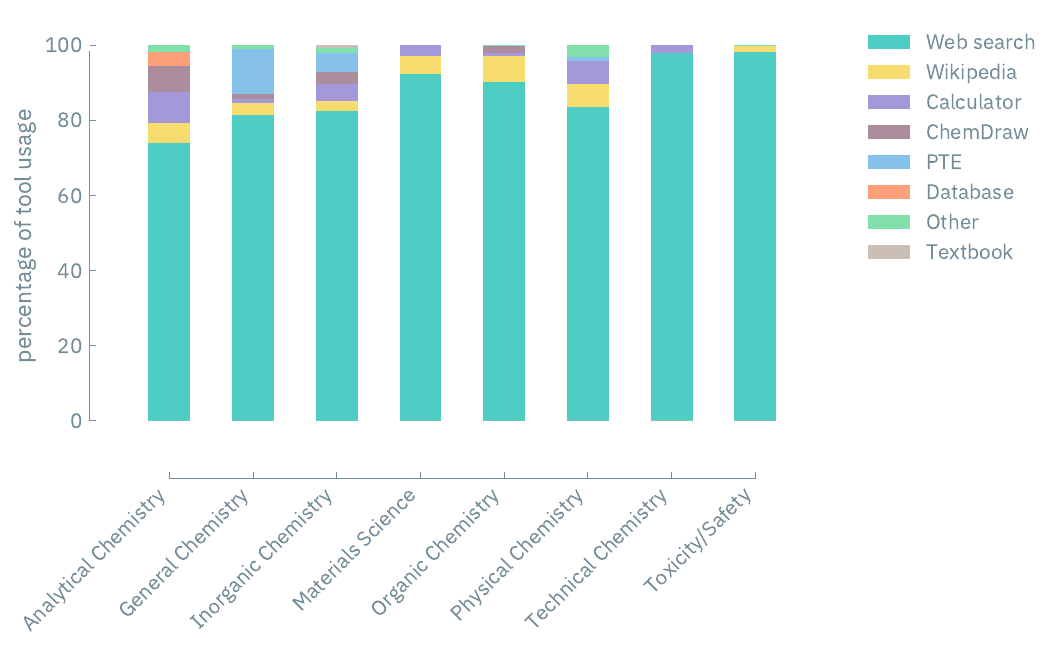}
    \script{human_tool_usage.py}
    \caption{\textbf{Tool usage by human scorers.} The plot shows the most commonly used tools by human participants.}
    \label{fig:tool_use}
\end{figure}

\paragraph{Tool use}
In our study, humans were allowed to use tools for answering some questions.
They could also report what tools they used for answering questions. As \Cref{fig:tool_use} shows, the most common tool was some form of web search (which, according to the free text responses, often was a multi-step process).

\clearpage
\subsection{Tool augmented models} \label{sec:react-environment}
In addition to directly prompting \glspl{llm}, we also investigated the performance of tool-augmented systems.
For this, we investigated the two models that showed the best overall results, \GPTFourO and \ClaudeThreeFiveSonnet (\oone overperformed both models but is not recommended using this model with such \enquote{reasoning} prompts such as ReAct\autocite{yao2022react} or \gls{cot}\autocite{wei2023cot}).
For both models, we created a ReAct-style tool augmentation environment in which models had access to WolframAlpha, the ArXiv \gls{api}, Wikipedia, and web search (using Brave search \gls{api}).
We based the selection of these tools on the most used tools by humans (see \Cref{fig:tool_use}).
Additionally, we added two specific tools to convert \gls{iupac} names to \gls{smiles}, and \gls{smiles} to \gls{iupac} names.
These conversion tools allow us to understand better how the agents perform for specific questions in this agent environment configuration.
We implemented the systems using Langchain\autocite{Chase_LangChain_2022} with the default ReAct prompt and constrained the system to a maximum of ten \gls{llm} calls.

While for \ClaudeThreeFiveSonnet the overall performance did not change, we observe a decrease in performance for \GPTFourO compared with the \gls{llm} without tools (\Cref{tab:performance_table}).
If we study the results by each type of question, we observe an improvement for questions regarding electron counts or point groups of compounds.
However, the scores decreased for questions about the number of isomers or \gls{ghs} pictograms.
For the specific questions about converting \gls{iupac} names to \gls{smiles}, the results decreased notably despite the models having access to specific tools prepared for those questions.
By studying the reasoning path for these cases, we found that the error results from the models responding in a format that the LangChain framework with the default ReAct loop cannot handle.
This indicates that agent frameworks need optimization to be more robust. It involves not only equipping the \glspl{llm} with tools but also necessitates engineering efforts to create robust systems.

\clearpage
\subsection{Confidence estimates} \label{sec:confidence_estimates}

Since it is important to understand if models can provide an indication of whether their answer might likely be incorrect, we prompted some of our top performing \glspl{llm} to return the confidence in providing a correct answer on an ordinal scale.
This is similar to the verbalized confidence scores reported by \textcite{xiong2023llms}.
We find that the models show different distributions of confidence scores, which, for some, are skewed to the extremes.

In addition, we also analyzed the confidence estimated via the log probabilities of the answer tokens. This probability of a token given the context is not necessarily the same as the confidence in the correctness of the answer. However, it is still often used as a proxy.

Our analysis of both log probabilities and prompting confidence reveals distinct calibration behaviors across different language models.
\GPTFourO demonstrates an overconfident tendency, often assigning high probabilities even to incorrect answers.
However, when \GPTFourO displays high confidence, it accurately predicts correct answers approximately \SI{80}{\percent} of the time. In contrast, \LlamaThreeOneEightBInstruct confidence distribution is more evenly spread, with a majority of predictions centered around 0.5. High-confidence predictions from \LlamaThreeOneEightBInstruct are less frequent compared to \GPTFourO, and unlike \GPTFourO, high confidence does not necessarily correlate with a higher chance of correct answers.

\begin{figure}[htb]
    \centering
    \includegraphics[width=\textwidth]{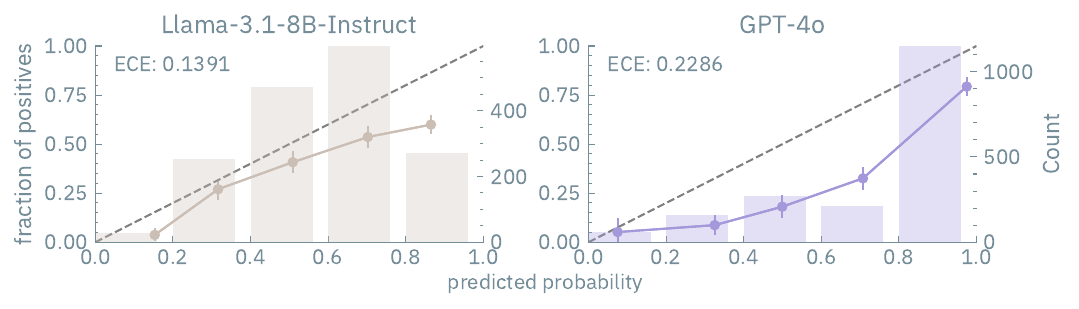}
    \caption{\textbf{Reliability diagram of logit-based confidence estimates.} For this analysis we obtained the linear probability from the logprobs of the models. Only logprobs of the tokens corresponding to the answers were considered. Linear probability was computed by taking exponential of logprobs (for sequences with multiple tokens, values were multiplied).  The plot shows the average predicted probability and the fraction of correct answers for each bin of linear probabilities. The ideal scenario is a diagonal line, indicating perfect calibration where the model's confidence aligns perfectly with the actual correctness. The \gls{ece} value quantifies the overall calibration performance, with a lower \gls{ece} indicating better calibration.}
    \label{fig:confidence_score_distributions}
    \script{plot_logprobs.py}
\end{figure}

\clearpage
\subsection{Impact of sampling temperature}
\begin{figure}[!h]
    \centering
    \includegraphics{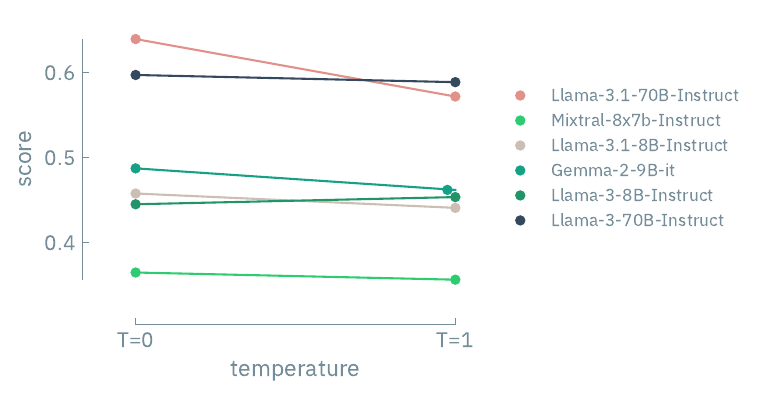}
    \caption{\textbf{Impact of sampling temperature on the performance of the models.} The plot shows the performance of the models at zero temperature (i.e., greedy decoding) and temperature of one. The performance is measured in terms of the fraction of questions answered correctly.}
    \label{fig:temperature_impact}
    \script{plot_temperature_diffs.py}
\end{figure}

We also investigated the impact of sampling temperature (i.e. temperature 0 means always sampling the most likely token, higher temperatures introduce some randomness in the generation process) on the performance of the models. \Cref{fig:temperature_impact} shows that, generally, the performance of models tends to decrease with increasing temperature.

\clearpage
\subsection{Summary of trends and recommendations}

\oone consistently leads across both the \chembench corpus and \chembenchmini.  \ClaudeThreeFiveSonnet, \GPTFourO, \LlamaThreeOneFourZeroFiveBInstruct form the next tier of strong performers. These models show robust performance across all skills and difficulty levels.

Larger models generally perform better (\LlamaThreeOneFourZeroFiveBInstruct > \LlamaThreeOneSeventyBInstruct > \LlamaThreeOneEightBInstruct). However, smaller but well-designed models can compete with larger models. For example, \GemmaTwoNineBIt has an overall accuracy of %
  0.48 \unskip\label{output/trends_section_variables/gemma_9B.txt}\unskip%
 on the \chembench corpus, which is only %
  10 \unskip\label{output/trends_section_variables/diff_between_llama_405B_and_gemma_9B.txt}\unskip%
\% less accurate than the much larger \LlamaThreeOneFourZeroFiveBInstruct model.

One can observe a clear progression of performance across model families (e.g., \ClaudeThreeFiveSonnet >  \ClaudeThree > \ClaudeTwo or \GPTFourO > \GPTFour > \GPTThreeFiveTurboZeroT). Newer versions consistently outperform their predecessors.
Scores on knowledge-intensive tasks are typically lower than those on calculation and reasoning-intensive questions.

Based on these findings, some scope for improving these models could be to focus on enhancing knowledge-based training, possibly through improved pre-training on chemistry-specific texts or with chemistry knowledge-base integration. However, innovation is also needed to incorporate this information into the systems. For example, even \PaperQATwo could not outperform the \gls{llm} it is used for directing the tools, even though the agent has access to literature evidence.
This suggests we might need to not only focus on building chemistry-specific retrieval datasets since current systems fail to retrieve the relevant papers and should be coupled with more domain-specific databases. However, the observation that our ReAct agents were too fragile to answer questions for which they had custom-made tools available correctly suggests that we also must invest in building more robust agent frameworks (see \Cref{sec:react-environment} for further discussion about the ReAct environment).

Models show the strongest performance in reasoning tasks on the \chembench corpus, which suggests that current training approaches are good at developing logical/reasoning capabilities for basic problem-solving of university-level exams. However, performance drops significantly for advanced tasks, suggesting that more advanced chemistry problems with complex, multi-step solutions (e.g., solving analytical chemistry problems) should be included in training or finetuning.

The post-training (e.g., \gls{rlhf}) of the current models does not equip them with human-like \enquote{chemical taste.} This, however, will be essential for future discovery research (e.g., in combination with genetic algorithms).

\clearpage
\subsection{Leaderboard}
\label{sec:leaderboard}
Our leaderboard is based on the tool chain developed for Matbench.\autocite{Dunn_2020}
Briefly, the \chembench pipeline produces standardized files in \texttt{json} format that contributors can add via pull requests to the \chembench repository.
The Markdown tables and interactive plots are automatically generated and updated on the \chembench website. The leaderboard is available at \url{https://lamalab-org.github.io/chem-bench/leaderboard/}.

\clearpage

\printnoidxglossary[type=\acronymtype, nonumberlist]  

\clearpage
\printbibliography[heading=subbibintoc]
\end{refsection}
\end{document}